\documentclass[11pt]{article}
\usepackage{multirow}
\usepackage[output-decimal-marker={,}]{siunitx}
\usepackage[a4paper,top=2.5cm,bottom=2.5cm,left=2.5cm,right=2.5cm]{geometry}
\usepackage{booktabs}
\usepackage{array}
\usepackage{rotating}
\usepackage[english]{babel}
\usepackage[T1]{fontenc}
\usepackage{hyperref}
\usepackage{longtable}
\usepackage{indentfirst}
\usepackage{amssymb, amsmath}
\usepackage{graphicx}
\usepackage{graphics}
\usepackage[table]{xcolor}
\newcommand{\R}{\mathbb{R}}

\usepackage{float}
\usepackage[normalem]{ulem}
\usepackage{soul}

\begin{document}
\title{Calibrating chemical multisensory devices for real world applications: An in-depth comparison of quantitative Machine Learning approaches}
\date{}
\author{S. De Vito$^{1,*}$, E. Esposito$^1$, M. Salvato$^1$, O. Popoola $^2$,\\
F. Formisano$^1$, R. Jones$^2$,  and G. Di Francia$^1$ \\
	$^1$ DTE--FSN--DIN, ENEA, P.le E. Fermi, 1, 80055 Portici (NA), Italy;\\
	$^2$ Chemistry Dept., University of Cambridge, United Kingdom\\
	$(^*)$ \texttt{Corresponding Author e-mail:\ saverio.devito@enea.it}}
\maketitle
\begin{abstract}
	Chemical multisensor devices need calibration algorithms to estimate gas concentrations. Their possible adoption as indicative air quality measurements devices poses new challenges due to the need to operate in continuous monitoring modes in uncontrolled environments. Several issues, including slow dynamics, continue to affect their real world performances. At the same time, the need for estimating pollutant concentrations on board the devices, especially for wearables and IoT deployments, is becoming highly desirable. In this framework, several calibration approaches have been proposed and tested on a variety of proprietary devices and datasets; still, no thorough comparison is available to researchers. This work attempts a benchmarking of the most promising calibration algorithms according to recent literature with a focus on machine learning approaches. We test the techniques against absolute and dynamic performances, generalization capabilities and computational/storage needs using three different datasets sharing continuous monitoring operation methodology. Our results can guide researchers and engineers in the choice of optimal strategy. They show that non-linear multivariate techniques yield reproducible results, outperforming linear approaches. Specifically, the Support Vector Regression method consistently shows good performances in all the considered scenarios. We highlight the enhanced suitability of shallow neural networks in a trade-off between performance and computational/storage needs. We confirm, on a much wider basis, the advantages of dynamic approaches with respect to static ones that only rely on instantaneous sensor array response. The latter have been shown to be best choice whenever prompt and precise response is needed.  
\end{abstract}

\section{Introduction}
Several research works have now highlighted the viability of low cost sensors based air quality monitoring systems (AQMS) for pervasive monitoring tasks \cite{eunetair1215}. Most of them are based on passive sampling, in which the sensors are freely exposed to the air to be analyzed. In these systems, real time sensing is hence continuously performed, without resorting to typical artificial olfaction measurement procedures. Eventually, the availability of fixed and mobile analyzers will probably lead to hybrid networks in which indicative low cost measurement systems will supplement the use of conventional analyzers. As the AQMS are based on low-cost technologies, it will then possible to solve the sparsity problem that negatively affect the current monitoring strategies \cite{lyons1990} \cite{alloway1997} \cite{bloomfield1996}. Some of these technologies could become available for citizen's in terms of wearable systems allowing them to obtain information on their personal pollutants exposure. Citizens involvement, as well as their contribution to the actual city wide measurement process, is foreseen on the basis of factual data by the results of several pilot projects \cite{progetto1,progetto2,progetto3}. Furthermore, this development will encourage good working relationship between citizens and public authorities, especially in those countries where this relationship is actually severely hampered \cite{Bianchi2004}. 
\newline\indent The European Commission has defined in \cite{EuDir} data quality objectives (DQO) for novel devices devised to be integrated in the operative air quality network. In that legal framework, these novel technologies are termed indicative measurement systems (IMS). In particular, their performance results needs to meet specific requirements including bias and uncertainty metrics. If met, the captured data could eventally be used in official air quality assessments and reports. In order to achieve quantitative pollution estimation capability, chemical sensors data needs to undergo a processing step, typically involving a regression algorithm, sometimes refered to as calibration function. Some sensors non-selective behaviour, relatively slow dynamic, instability and sensitivity to environmental conditions, severely hinder the use of raw sensor data as estimations for real world measurements. To this regard, the Aveiro intercomparison exercise have set a a valuable point of reference evaluating the actual field performance of several multisensors, based on different classes of sensors, targeted to multiple pollutants and comparing their performances with EU DQOs \cite{Aveiro}. This achievement required the involvment of several research groups across Europe and multiple large, medium and small enterprises. Specifically, chemical microsensors devices are, in general, subjected to interferent gases that modify their response to the target gas \cite{marco1996}. For this reason, any attempt to rely on a univariate calibration procedure, neglecting interferents influence, is prone to failure \cite{devito2009}. Any available information on interferent gases concentration should be exploited in order to solve this issue, hence advocating the use of multivariate calibration algorithms. Combined with sensors non linearity, cross-sensitivity have suggested the use of machine learning algorithms to solve the calibration problem. Besides, chemical microsensors response generally changes in time due to several effects including poisoning and environmental variables sensitivity \cite{rock2008electronic}. As a consequence, long term stability is a significant concern given the need to limit recalibration and maintenance burden on a pervasive network of possibly hundreds of AQMS \cite{marco2012signal}. Adaptive/semi-supervised calibration schemes could represent one of the possible solutions to measurement drift issue \cite{Martinelli2013}. Furthermore, owing to method of fabrications, some sensors have been shown to have poor sensor-to-sensor reproducibility. This hampers the use of a single calibration function, thus requiring an ad hoc calibration procedure for each chemical multisensor device or, alternatively, the development of calibration transfer strategies \cite{calTransfer1}, \cite{calTransfer2}. As such, calibration transfer is an active and relevant field for researchers in artificial olfaction. A significant research effort is also needed to further develop our chemical sensor data fusion capabilities, in particular, to precisely reconstruct a 3D picture of pollution in both indoor or outdoor \cite{3D},\cite{Capelli2014},\cite{Achim2013}. 
\newline\indent The possibility to quantitatively assess our own exposure to pollutants when moving in the city, is currently a strong and desiderable objective with the added advantage of possible impact on citizen's mobility, active life and health consciousness. While these are desirable, it may require the execution of such intelligent data processing algorithms on board of the citizen's mobile or pervasively distributed fixed nodes. This will make the resulting monitoring network a true smart cyber chemical system. In fact, scalability issues, that particularly affect networking performances in a more and more crowded IoT scenario, will require the local execution of data preprocessing without resorting to the cloud. On these basis, the emerging ''fog'' and ''edge'' computing frameworks are also pushing low semantic extraction computations towards the very edge of a sensing or control network \cite{bonomi2012fog}. Besides, such a monitoring network will be based on heterogeneous sensing systems and the possibility of obtaining local precise and accurate estimations of pollutants concentrations will be a mandatory requirement to achieve the needed plug and sense capability. 
Several algorithms have been proposed during the last decade to implement the needed quantitative calibration of the chemical multisensor device but currently no effective comparison have been performed \cite{marco2012signal,GutierrezOsuna2002,marco1996}. The main reason is the lack of publicly available datasets to assess performance of the different approaches. As such, many questions are still open, such like: which approach best perform under certain conditions, which one is the most efficient or in other words which one need less samples to produce satisfactory results. Other questions include: which one is the most computationally efficient and which one has the most efficient knowledge representation. These, are perceived by many as the most interesting in order to advance towards the development of smart cyber chemical systems. Very recently, the availability of low-cost technologies and open source hardware has facilitated the development of chemical multi-sensors prototypes. Field tests are generating a consistent flow of interesting data. Simultaneously, a new generation of researchers are actively sharing the data they gather both in laboratories and real world deployment making it possible to build comprehensive comparisons. In this work, we perform a comprehensive benchmarking of the most promising techniques that have shown reliability in the depicted scenarios at least in laboratory experiments, focusing on machine learning approaches. The use of multiple datasets (two of which are publicly available) will provide the needed robust test of our analysis. The presented techniques will be assessed for their ability to be implemented on-board the devices as well as how well they could meet EU requirements for indicative methods in terms of accuracy and precision level. In section two, we will review the multivariate regression system,s recently proposed in literature for chemical sensor data analysis, selecting the most promising ones according to the results obtained by researchers in the field. Section three will define the benchmarking scenarios chosen to estimate algorithms performances in this challenging framework. This include performance drivers, dataset description and performance indicators selection. Our results and conclusions are covered in the last two sections. 

\section{Related works}
Historically, quantitative calibration of chemical multi-sensor devices relied on univariate calibration and simple environmental factors correction strategies. Sensor responses captured during laboratory calibrations were used to obtain a linear relationship among the sensor response and its target gas concentration. The limited performance obtained by this strategy has led to the development of multivariate calibration strategies relying on the use of complex synthetic gas mixtures \cite{zhang2013,althainz1996} or on field recorded data \cite{kamionka2006} to cope with specificity and stability problems. In the first case, researchers usually relied on the use of steady state responses for the calibration computation. This usually requires a significant amount of time in order to explore a wide experimental space generated by the possible combinations of different concentrations of several gases. Furthermore, most of the recent systems operate in open sampling scenarios in which sensors rarely reach a steady state \cite{trincavelli2009}. Dealing with the use of mobile chemical sensing systems, Hishida et al. \cite{ishida1998}, for example, explicitly proposed the use of steady state responses as approximates of the real sensors responses in field deployments. Conversely, on field recorded data made it possible to obtain convenient and efficient calibration datasets. In fact, these datasets cover the actual manifold (trajectory) of gas concentrations values that a multi-sensor will likely encounter when deployed in the field. Simultaneously, machine learning experts have proposed data intensive approaches that simplified the need to develop apriori and complex non-linear models based on the actual physics of the devices response towards their target gas and interfering species. A recent example of a practical, physically rooted, model has been provided by Masson et al. The model can be calibrated with laboratory based sensor response recordings and tuned with on field recorded data to compensate for temperature interference \cite{masson2015}. However, the approach should be improved to address multiple interferents. Furthermore, to the best of our knowledge no such class of solution has yet been proposed to simultaneously model the sensor dynamic behaviour. Machine learning approaches, instead, rely on a quasi-black box approach in which the knowledge on the sensor model can be used for response descriptive features development. The actual model selection is performed automatically, by tuning a generic nonlinear model, allowing it to learn using example in a typical supervised fashion. In the last few years, multiple methodologies have been proposed relying on the most common and efficient machine learning regression strategies like shallow neural networks, support vector machines, gaussian processes and more recently, reservoir computing. The last has the advantage of implicitly learning a dynamic model of the multisensor device, thereby reducing the effects on the performance of the slow sensors dynamic \cite{fonollosa2015}. These effects can be paramount when dealing with pervasive and mobile deployments where systems are likely to encounter rapid concentrations transients. Advanced learning techniques like semi-supervised learning proved to be promising in reducing the number of calibration samples needed as well as improving robustness to drift by complementing learning from un-labeled samples with the classic supervised approach \cite{Martinelli2013}.
Shallow artificial neural networks (ANN) have been proposed by multiple researchers. De Vito et al., for example, have shown promising results using hourly averaged sensors  data from field measurements; this work showed the possibility to let them learn a nonlinear multiple regression model for benzene concentration estimation \cite{devito2008field}. Further works from the same group have evaluated neural networks generalization properties in terms of performance against number of training samples, suggesting the possibility of using feature selection to improve performance, resulting in better insights on the actual cross sensitivities and correlation behaviour. The effects of the changes of pollutants joint concentrations distribution, which can be due on seasonal effects, on the performances were also highlighted \cite{devito2009}.
Spinelle et al., have confirmed that shallow neural networks, can significantly outperform linear univariate and multivariate models, thus highlighting the possibility to exploit multivariate information to reach EU set DQOs \cite{spinelle2015}. Very recently, Al Barekh et al. \cite{albarakeh} proposed a fuzzy logic based strategy for differentiating among different type of air pollution and estimating a pollution index using a neural network. The resulting algorithm was used for calibrating a small fleet of three open sampling chemical multi-sensor devices. Support Vectors Machines (SVMs) have been proposed for gas mixture detection, identification and quantification problems using multi-sensor systems by several authors. Unfortunately, only a few works are related with open sampling systems. Kai Song et al. \cite{kay_song} used a least square SVM for calibrating a wireless chemical multi-sensors system which targeted explosive gases (methane, hydrogen) concentration estimations. Their on-board implemented (Least Squares)-SVM solution outperformed an ANN solution on a small scale validation experiment with near low explosion limit concentrations. In order to cope with sensors dynamic behaviour, Vembu et al. \cite{vembu2012} proposed the use of time series kernel based SVMs for enhancing the identification capabilities of a small network of pervasive open sampling multi-sensors systems. In this work, authors reported an enhanced accuracy performance with respect to basic Radial Basis Function (RBF) kernels SVM working on time series features. In an effort to improve the response of an open sampling system while exposed to rapid concentration changes, Fonollosa et al. \cite{fonollosa2015}, \cite{SHEIK2014}, investigated the use of reservoir computing strategies. In this approach, a recurring neural network using a large reservoir of randomly connected nonlinear computational units and a layered linear regressor system was used to train the model. The ultimate goal was to precisely estimate the actual concentration of the stimulus gases in a pseudo-controlled setting and the authors reported superior performances of their Reservoir Computing (RC) system with respect to static SVM and linear approaches. Interestingly enough, the random wiring of the recurrent units of this architecture allowed to save the time needed for selecting the optimal sensors time serie window of observation that is paramount for the  dynamic modelling problem. Esposito et al. \cite{esposito2016dynamic}, investigated the performance of the tapped delay approach with real world data, recorded with open sampling systems. They highlight that dynamic approaches may be well suited for tackling rapid transient exposure encountered in the field by mobile systems or fixed roadside stations. Robotics olfaction is a closely related field that will definitely benefit from improved real time gas quantification capability. Monroy et al. described an interesting probabilistic quantification approach based on the well-known Gaussian processes (GP) framework \cite{monroy2013}. In their first work \cite{monroy2012}, instead of attempting to describe sensors dynamics, they propose the prediction of the uncertainty caused by slow sensors operation in an open sampling scenario. They validate the approach within a simulated field environment, in which ethanol is emitted and transported by an air flow towards MOX (Metal Oxide) sensors located in a room. A photo-ionization device (PID) VOC sensor was used as a reference. They reported similar performance with respect to state of the art static SVRs. In follow up work, they developed the use of GPs towards a dynamic approach, coupling a tapped delay line and computing sensor derivatives. Their results showed a slight improvement of accuracy performance indicators in the case of derivatives features.

\section{Benchmarking scenarios}
\begin{figure}[H]
 	\centering
 	\includegraphics[height=250pt,width=400pt]{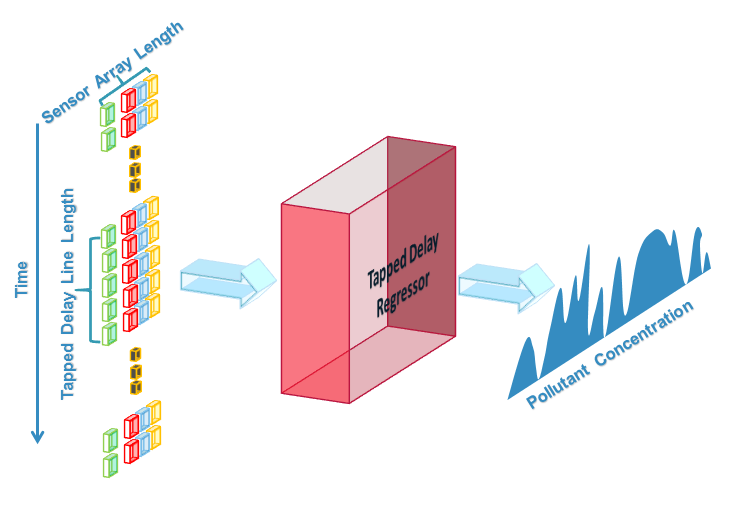}
 	\caption{Schematic rendering of the tapped delay architecture used for the dynamic approaches. The response of the relevant sensors observed during a pre-set length and moving time window, is fed to the machine learning component in order to promptly estimate present concentration irrespective of fast transients.}
 	\label{architecturetdl}
 \end{figure}

In the previous sections, we briefly reviewed relevant works that have addressed the chemical multi-sensor arrays calibration problem with machine learning approaches. As mentioned above, no comprehensive comparison of the proposed architectures is currently available, partially due to the lack of public datasets to work with. In this section, we introduce the comparison methodology that we have pursued. It is based on the availability of three different datasets. A selection of the most promising machine learning techniques have been trained alongside with their dynamic extension to estimate the concentration of different pollutants by using instantaneous sensor responses and, for dynamic architectures, their recent trajectories. Specifically, a small window of observation including recent past sensor responses is, in these cases, used to train the model thereby taking into account the sensor dynamic behaviour (see Fig. \ref{architecturetdl} ). 
\newline\indent A small set of literature based performance indicators has been chosen to represent the performance of the architectures under comparison. Since the ultimate goal of this systems is to be operated on-board or in high efficiency big data processing facilities, we briefly review the complexity of these methods with special focus on their operative phase, i.e. the computational and storage burden of the concentration estimation procedure. 
\subsection{Machine Learning Techniques }
The Machine Learning (ML) techniques that we have taken into account for comparison purposes are among the most popular including Neural Networks (NN), Support Vector Regressors (SVR), Gaussian Processes Regressors (GPR), Multiple Linear Regressors (MLR) and Reservoir Computing (RC) along with their dynamic versions, obtained using different Tapped Delay Lengths (TDL) at the input. The tapped delay allows for a time domain expansion of the inputs generating an observation window for the machine learning technique that allows for learning and exploiting a knowledge on dynamic relationships between input (sensor responses) and outputs (pollutant actual concentrations). 
The considered techniques have been analysed for their capability to provide a solution to the need of on-board implementation as well as how well they meet EU requirements for indicative methods in terms of accuracy and precision level. 
Neural networks have been used in this scenario since the last decade  \cite{devito2007,pardo2000}. Their basic architecture is characterized by multiple layers of information-processing nonlinear functional units interacting by means of one way weighted connections. Mathematically, their prediction equation may be written as in the following: 
\begin{equation}\label{NN}
\begin{aligned}
	u_k & = \sum_{j=1}^{n}w_{kj}\cdot x_j \\ 
	y_k & = \varphi(u_k+b_k),
\end{aligned}
\end{equation}
where $x_j,\, j=1,\ldots,n$ are the input vectors, $u_k$ is the weighted sum of inputs, $b_k$ is the bias term, $\varphi(X)$ is the activation function (e. g. \texttt{sigmoid functions, hyperbolic tangent, sign function, etc.}) and $y_k$ is the output of the single NN neuron $k$. Usually this structural pattern is replicated multiple times, one time for each layer, until the final one outputtinig the prediction. 
\indent Support Vector Regressors (SVR), are usually selected for their potential enhanced performance and computational advantages. Their regression function can be actually computed by means of a small subset of training points called the support vectors.  Practically, in SVR, the input $\mathbf{X}$ is first mapped onto a m-dimensional feature space using some fixed (nonlinear) mapping, and then a linear model is constructed in this feature space. Using mathematical notation, the linear model (in the feature space) $f(\mathbf{X},\omega)$ is given by
\begin{equation*}
f(\mathbf{X},\omega) = \sum_{j=1}^{m}\omega_j g_j(\mathbf{X})+b,
\end{equation*} 
where $g_j(\mathbf{X}), j=1,\dots,m$ denotes a set of nonlinear transformations, and $b$ the 
\texttt{bias} term.
 In the primal optimization problem, we would like to minimize the sum of the empirical estimation errors on the training set samples in a trade off with a solution complexity term,  $ \omega ^{T} \omega $,  regulated by a constant C.  
Using an epsilon insensitive loss function, i.e. by disregarding errors not exceeding a fixed term epsilon,  we ensure existence of the global minimum and, at the same time, the optimization of reliable generalization bound.
It is well known that SVR generalization performance (estimation accuracy) depends on a good setting of hyperparameters $C$, $\epsilon$ and the kernel parameters\footnotemark[1]. The problem of optimal parameter selection is further complicated by the fact that SVR model complexity (and hence its generalization performance) depends simultaneously on all three parameters, 

\footnotetext[1]{The solution of the dual problem is given by:
	\begin{equation*}
	f(\mathbf{X}) = \sum_{i=1}^{n_{sV}}(\alpha_i - \alpha^{*}_{i})K(\mathbf{X_i},\mathbf{X})+b,
	\end{equation*} 
s.t. $0\leq\alpha^{*}_{i}\leq C, 0\leq\alpha_{i}\leq C$, where $n_{sV}$ is the number of Support Vectors ($SV_s$) 
and the kernel function $$K(\mathbf{X},\mathbf{X_i})=\sum_{j=1}^{m}g_j(\mathbf{X})g_j(\mathbf{X_i}).$$}
\indent GP take a nonparametric approach to regression and offers a stochastic regression function. It is completely characterized by its \textit{mean} and \textit{covariance function} or \textit{kernel}.
Consider the training set $ \left\lbrace (x_i,y_i); i=1,2,\dots,n \right\rbrace $, where $x_i\in \R^d $ and $y_i\in\R$, an istance of response $y$ from a GPR model can be modeled as $$P\left(y_i|f(x_i),x_i\right) \sim N\left( y_i|h(x_i)^T \beta +f(x_i),\sigma^2\right).$$ 
Hence, a GPR model is a probabilistic model. There is a latent variable $f(x_i)$ introduced for each observation $x_i$, which makes the GPR model nonparametric. In vector form, this model is equivalent to $$P(y|f,X)\sim N(y|H\beta+f,\sigma^2 I),$$ where
\[X=
\begin{pmatrix}
x^{T}_{1} \\
x^{T}_{2} \\
\vdots \\
x^{T}_{n}
\end{pmatrix}, \quad 
y=
\begin{pmatrix}
y_1 \\
y_2 \\
\vdots \\
y_n
\end{pmatrix}, \quad
H=
\begin{pmatrix}
h(x^{T}_{1})\\
h(x^{T}_{2})\\
\vdots \\
h(x^{T}_{n})\\
\end{pmatrix}, \quad
f=
\begin{pmatrix}
f(x_1) \\
f(x_2) \\
\vdots \\
f(x_n) \\
\end{pmatrix}.
\]
The joint distribution of $f(x_1),\ldots,f(x_n)$ is $$P(f|X)\sim N(f|0,K(X,X)),$$
where $K(X,X)$ looks as follows:
\[
K(X,X)=
\begin{pmatrix}
k(x_1,x_1) & k(x_1,x_2) & \cdots & k(x_1,x_n) \\
k(x_2,x_1) & k(x_2,x_2) & \cdots & k(x_2,x_n) \\
\vdots     & \vdots     & \vdots & \vdots     \\
k(x_n,x_1) & k(x_n,x_2) & \cdots & k(x_n,x_n) \\
\end{pmatrix}
\]
The covariance function $k(x,x^{'})$ is usually parametrized by a set of kernel hyperparameters $\theta$. 
In particular, we have used the Matlab function \texttt{fitrgp}, which estimates the noise variance $\sigma^2$ and the hyperparameters $\theta$ of the kernel function from the data while training the GPR model.
\newline\indent MLR regressors are used as a state of the art systems for comparison reasons. Assuming that $\textbf{X}$ is the input features, the classic mathematical formulation of MLR model is $$ y=\textbf{X}\beta+\epsilon,$$ with $\epsilon$ is the intercept and $y$ the predicted value.
\newline\indent In this work we foster the use the Reservoir computing approach, originating from the echo-state network paradigm and introduced in chemical multisensors field by  \cite{fonollosa2015}. This two-stages approach is based on a network structure that is characterized by multiple recurrently interconnected non linear units, called the reservoir,  that realize a time-expansion of the input sequence. After this expansion, the so processed inputs are fed in a MLR based final processing layer that eventually output the prediction. An interesting feature of this algorithm is that the recurrent non-linear dynamic units, more properly their weighted interconnections, are not trained,  only the output linear layer is. Actually, the non-linear elements in the reservoir are randomly interconnected at the start of the training phase actually determining the expansion behavior that will remain fixed. More specifically the random interconnection process is regulated by two parameters namely $input \; scaling$, chosen such that the input activity can induce sufficient activity in the reservoir and $spectral \; radius$, chosen to ensure that the resultant activity of the reservoir, when the inputs occurr, is sufficiently diverse for different inputs. The time behavior of the RC model,  is mathematically defined by: 
\begin{align}
\mathbf{\tilde{x}(n)} & =\tanh(\mathbf{W}^{in}[1; \mathbf{u}(n)]+\mathbf{Wx}(n-1)),\\
\mathbf{x}(n) & = (1-\alpha)\mathbf{x}(n-1)+\alpha\mathbf{\tilde{x}(n)},
\end{align}
where $n$ is discrete time, $\mathbf{u}(n)\in\mathbb{R}^{N_u}$ is the input signal, $\mathbf{x}(n)\in\mathbb{R}^{N_x}$ is a vector of reservoir neuron activations and $\mathbf{\tilde{x}(n)}\in\mathbb{R}^{N_x}$ is its update at time step $n$. $\mathbf{W}^{in}\in \mathbb{R}^{N_x\times N_u}$ is the input weighted matrix and $\mathbf{W}\in\mathbb{R}^{N_x\times N_x}$ is the recurrent weighted matrix; $\alpha\in\left( 0,1\right]$ is the leaking rate. we assume $\alpha=1$, and thus $\mathbf{\tilde{x}(n)}\equiv\mathbf{x}(n)$.
The linear readout is defined as 
\begin{equation} \label{readout}
 \mathbf{y}(n)=\mathbf{W}^{out}[1;\mathbf{u}(n);\mathbf{x}(n)],
\end{equation}
where $\mathbf{y}(n)\in\mathbb{R}^{N_y}$ is the network output and $\mathbf{W}^{out}\in\mathbb{R}^{N_y}$ the output weighted matrix.
we used the open-source Python library Oger \cite{oger2012} for the implementation of the algorithms developed to compute gas concentration estimations.

\subsubsection{Computational Costs} 
The working principle of Machine Learning algorithms encompasses two main steps that is a training phase, in which learning by example take place, and an operative phase, during which the trained system process input data and performs its estimations. Training phase is usually far more computationally intensive that a single estimation operation, usually requiring the evaluation of hundreds or thousands of sensory data instances and their \uline{respective} reference data in order to tune the machine learning algorithm parameters.  However, in most applications, training is performed just once for the entire operative lifespan of the intelligent device.  In our specific case, machine learning algorithms are usually first trained using sensor responses to different air quality conditions for which reference data about pollutant concentrations are known. This phase usually is carried out off-line by dedicated systems. Afterwise, operative life is started and the tuned machine learning algorithm is continuously fed with sensor responses for carrying out estimations of pollutant concentrations. Some authors highlighted the possibility to retrain systems in order to correct for sensors and concept drift  but even in this case, training operation is performed very rarely, possibly once every sixth months. Smart cyber chemical systems are built by networks of chemical multisensory devices relying on microcontrollers with limited computing and storage capabilities. Their field recorded data are sent to centralized cloud computing facilities devised to process the incoming stream of sensory data coming from all the networked devices. In both cases, computational and storage resources of both subsystems are challenged by the amount of data to be processed in real time using complex machine learning algorithms. Hence, the computational and storage complexity of such algorithms is highly relevant and can determine the choice of the actual machine learning algorithm to implement. Due to the rare need to execute a (re) training phase, the operational phase computational and storage complexities are much more relevant, for our scenario, than those of the training phase.
\newline\indent The operational computational cost of a NN can be split into two main contributions. The first one is due to the linear part of the NN prediction equation, deriving from the operations needed to perform the sum of the weighted inputs of each neuron. The second one, is related to the computation of the neurons activation function. Considering eq. \ref{NN}, we have $k$ function invocations for each hidden layer being  $k$  the number of hidden neurons in the layer. Furthermore we have $k$ dot $n$ products coming from matrix products (where $n$ is the dimension of input vectors space). Storage and computational complexity is hence related with the total number of interconnection weights in each layer. Ultimately, for shallow neural networks including a single hidden layer and a single output, it scales like $O(nk)$.  \newline \indent Conversely, as we shown in the previous section, prediction complexity of kernel SVR depends on the choice of kernel and it is typically proportional to the number of support vectors. For most kernels, including polynomial and RBF, computation and storage complexity  is bound by $O(n_{SV}d)$, where $nSV$ is the number of support vectors and d is the dimension of the input space. 
\newline\indent GPR model, in order to provide predictions for new data, requires:
\begin{itemize}
	\item Knowledge of the coefficient vector $\beta$ of fixed basis functions;
	\item Evaluation of the covariance function $k(x,x^{'}|\theta)$, given the kernel parameters $\theta$;
	\item Knowledge of the variance $\sigma^2$, that appears in the density $P(y_i |f(x_i),x_i)$.
\end{itemize}
Training a GPR model requires the inversion of an $n+n$ kernel matrix $k(X,X)$. The memory requirement for this step scales as $O(n^2)$ since $k(X,X)$ must be stored in memory. One evaluation of $\log(y|X)$ scales as $O(n^3)$. Therefore, the computational cost is $O(kn^3)$, where $k$ is the number of function evaluations needed for maximization and $n$ is the number of observations.
The computation of the prediction values involves the estimation of $\alpha=(K(X,X|\theta)+\sigma^2 I_n )^{(-1)} (y-H\beta)$, which has the computational complexity equal to $O(n^3)$ and the memory requirement is $O(n^2)$.
For large $n$, estimation of parameters or computing predictions can be very expensive. The approximation methods usually involve rearranging the computation so as to avoid the inversion of an $n\times n$ matrix.
\newline \indent MLR has been trained using the Matlab function \texttt{fitlm}, in which we have selected the standard model, i.e. the model containing an intercept and linear terms for each predictor. As know, the computational cost of this algorithm depends on the size of the input matrix, in particular the matrix product $\mathbf{X}\beta$.  
\newline\indent  As we can see in eq. \ref{readout}, the estimation of the RC computing prediction $y_{target}(n)$, involves the computation of outputs weights via Linear Regression. In particular, the vectors $[1; u(n); x(n)]$ are collected in a matrix $\mathbf{X}$ and $y(n)$ in a matrix $\mathbf{Y}$, both having a column for every training time step $n$. The computation of the expansions $\mathbf{x}(n)$ costs $N_x$ function invocations, $N_x\cdot N_u$ products for the computation of $W^{in}[1;\mathbf{u}(n)]$ and $N_{x}^{2}$ products for the computation of $W\mathbf{x}(n-1)$. Since, typically, $N_x\gg N_u$, the computational complexity is $O(N^{2}_{x})$, the storage complexity being bound by the same term. 
These insights will be useful for completing a fair comparison of the tested techniques in section 4.

\subsection{Datasets Description} 
\indent The comparison was carried out taking into account three different datasets: the first was from a specific deployment of multi-sensory devices (SNAQ systems, see \cite{mead2013} for detailed description of the deployment), developed by the Centre for Atmospheric Sciences (CAS), Chemistry Department, University of Cambridge (UK). The second dataset was from a field deployment of multisensory device located in significantly polluted roadside within an Italian city. Finally, the third dataset was collected in a gas delivery platform facility at the ChemoSignals Laboratory in the BioCircuits Institute, University of California UCSD, USA. The measurement system platform provides versatility for obtaining the desired concentrations of the chemical substances of interest with high accuracy and in a highly reproducible manner. 

\subsubsection{SNAQ Dataset}
The passive multisensory device was equipped with the following sensors array: 
\begin{itemize}
	\item two different $NO_2$ electrochemical (EC) sensor units (Alphasense $NO_2-B4$ termed in the following as $NO_2(A)$ and $NO_2(B))$;
	\item one $NO$ Alphasense EC sensor unit (Alphasense $NO-B4$);
	\item two different $O_3$ Alphasense EC sensor units (Alphasense $O_3-B4$ termed in the following as $O_3(A)$ and $O_3(B)$);
	\item temperature and relative humidity (RH) sensor units;
	\item wind speed and wind direction device.
\end{itemize}	
Multiple instances of the multisensor were deployed within the city Centre of the city of Cambridge (UK) as a part of a pervasive deployment. One of them was located on the roof of the Chemical Dept. together with a conventional reference station operated by CAS.
A couple of the SNAQ units were deployed within Cambridge city centre (UK), as a part of a pervasive deployment. One of them was located on the roof of the Chemical Department, University of Cambridge. together with a conventional reference station operated by CAS. This station relied on certified chemiluminescence and spectrometer based analysers. While the SNAQ units samples at 20s interval, the reference station had a temporal resolution of 1 minute. This reference station monitored toxic gases including  carbon monoxide ($CO$), nitric oxide ($NO$),nitrogen dioxide ($NO_2$), ozone ($O_3$), $SO_2$.In this work, we considered all raw instantaneous sensors readings for calibration purposes, comparing the estimations results with the conventional analyzer samples when available (one out of three sensor readings). Here, it is worth to note that electrochemical sensors, when operating at low $ppb$ levels, are also prone to interference issues that limits the performance outcome. Actually a known cross sensitivity has been reported for, $O_3$ and $NO_2$ \cite{o3} and \cite{no2}. Together with temperature interference this effect is expected to represent the main limit to absolute performance in this dataset. 
Sensors were calibrated by the manifacturer with a linear univariate procedure, so raw sensors signals units were $ppb$. Baseline and temperature correction by using datasheet procedure were also implemented. Finally, the reference data was processed for unusable data period when daily calibration was carried out. These rejected data spans few minutes every day. In order to build a suitable dataset, five weeks of continuous measurements were used in this study.

\subsubsection{ENEA Pirelli Dataset}
The second dataset came from the source:  \url{https://archive.ics.uci.edu/ml/datasets/Air+Quality}. It contains 9358 instances of hourly averaged responses from an array of 5 metal oxide chemical sensors embedded in an air quality chemical multi-sensor Device. Data were recorded from March 2004 to February 2005 (one year) representing the longest freely available field deployed air quality data from this type of chemical sensor devices. Reference hourly averaged concentrations for CO, non methane hydrocarbons (NMHC), benzene, total Nitrogen Oxides ($NO_x$) and $NO_2$ were obtained from reference certified analyser where the multi-sensors were co-located. Evidences of cross-sensitivities as well as both concept\footnotemark[2] and sensor drifts are present as described in De Vito et al. \cite{devito2008field}. These  eventually represent the main drivers that affect sensors concentration estimation capabilities in the long term, determining the performance limits. The sensors array is equipped in the following way: 
\begin{enumerate}
	\item $PT08.S1$ (tin oxide) hourly averaged sensor response (CO species);
	\item $PT08.S2$ (titanium) hourly averaged sensor response (NMHC species);
	\item $PT08.S3$ (tungsten oxide) hourly averaged sensor response ($NO_x$ species);
	\item $PT08.S4$ (tungsten oxide) hourly averaged sensor response ($NO_2$ species);
	\item $PT08.S5$ (indium oxide) hourly averaged sensor response ($O_3$ species);
	\item Temperature, RH and Absolute Humidity.
\end{enumerate}
For all the sensor array, recorded signal unit were ohms. Similarly to the previous dataset, the device was operated in continuous operation mode but the timeframe differs considerably in that only hourly averages of both sensor recordings and reference data are available. 
\footnote[2]{Here we refer to Concept drift as the usually slow variation of process relevant variables distributions. In our case, these are the target and non target pollutants concentrations as well as RH and T.}
\subsubsection{UCSD dynamic gas mixtures Dataset}
The final dataset (available on \url{https://archive.ics.uci.edu/ml/datasets/Gas+sen}\\ \url{sor+array+under+dynamic+gas+mixtures}) contains the data from laboratory tests in which 16 chemical sensors were exposed to gas mixtures at varying concentration levels for 12 hours withou interruption. The sensor array was placed in a $60 ml$ measurement chamber, where the gas sample was injected at a constant flow of $300 ml/min$. 

In particular, two gas mixtures were generated: ethylene and methane in air, and ethylene and CO in air. The latter has been used in this work with concentrations designed to elicit significant interference in the involved sensors array.  Specifically, the sensor array included 16 MOX chemical sensors (Figaro Inc., US) made up of 4 different types: TGS-2600, TGS-2602, TGS-2610, TGS-2620 (4 units of each type).
  They were integrated with custom built signal conditioning and control electronics. The operating voltage of the sensors, which controls the sensors operating temperature, was kept constant at 5 V for the whole duration of the experiments. The sensors conductivities were acquired continuously at a sampling frequency of 100Hz. However, this was subsequently averaged to create 1s data. No drift correction procedure was implemented. The targeted reference concentration levels (set-points) were changed randomly each 80 to 120 seconds, and abruptly,  involving significant concentration changes (e.g. from 0 to 300 ppm of CO).  As above mentioned, the relatively high flow rate of the carrier gas ($300ml∕min$) allowed for fast exchange in the 60ml sensors chamber ($\sim 12s$ required to fill the chamber), reducing (but not nullifying) the dead volume inertia influence. However, no correction to target gas concentrations have been performed by taking into account sensor chamber dynamics. This challenged the relatively slow dynamics of the sensors. Fast and random  target concentrations (set points) transients make this dataset differ from the previous ones in which the true concentration levels varied smoothly in uncontrolled way. Indeed, the dataset was purposedly designed to induce significant transient and cross-interference errors that are the main drivers for limited absolute performances. The UCSD dataset was also built such that several gas mixtures compositions, included pure gases, were considered. 

\subsection{Performances analysis}
The above mentioned methodologies have been tested for their ability to model and create a generalised relationship between sensors response and the target gas concentration. The three datasets have been subset into training, validation and test sets by keeping the natural timing sequence i.e. by selecting sensor responses measured at subsequent times. The validation subsets were used for model selection i.e. for selecting the best performing hyper-parameters value set and tapped delay length. The test subset  were used to evaluate the predictive performance. We have also varied training set length to investigate the response of the different methodologies in terms of generalization capabilities. We optimised by conducting extensive exploration of model hyper-parameters subspaces for all architectures. Mean absolute error (MAE) defined as the sample mean of absolute prediction error and its standard deviation (STD) were used to control the hyper-parameter selection and reported as performance indicators. For NN, GPR and RC approaches, each training and test procedure were repeated 30 times to reduce the uncertainty induced in performance indicators by the random choice of booting parameters, respectively: NN initial weights, results of kernel hyperparameters selection and random reservoir units wiring. Best performing architectures for each ML technique, defined by their hyper-parameters t-uple, have been selected to be compared. 

In Table \ref{tabella:spaceParam} the explored space of hyperparameters is reported.
\begin{table}[H] 
		\resizebox{1\columnwidth}{!}{ \begin{tabular}{ccccc} 
			\toprule
		     {NN} & {SVR} & {GPR} & {MLR} & {RC}  \\
			\midrule
			{Hidden Neurons Number (HNN)} & {Kernel Function (KF)} & {Kernel Function} & {-} & {Spectral Radius $(\rho)$} \\
			{Epochs Number} & {Kernel Scale $(\gamma)$} & {Initial value for the noise std of GP model $(\sigma)$} & {-} & {Input Scaling (IS)} \\
			& {Box Constraint (C)} &  & {-} & {Reservoir Units (RU)} \\
			& {Epsilon-insensitive band $(\epsilon)$} &  &  & \\
			\bottomrule 
		\end{tabular}}
		\caption{Hyperparameters space.}    
		\label{tabella:spaceParam}
	\end{table}
For NN, the hyperparameters that we considered were:
\begin{itemize}
	\item Hidden neurons number (HNN),
	\item Epochs number (EN),
\end{itemize}
for each tapped delay length.
In particular,  $HNN$ varied in the [3, 5, 7, 10, 15, 20] set while $EN\in [100, 200, 300, 400, 500, 600, 900]$.
\newline For SVR we investigated the hyperparameters space consisting of:
\begin{itemize}
	\item Kernel Scale factor $\gamma\in\left(2^{-15}, 2^5 \right)$;
	\item Box Constraint $C\in\left(2^{-5}, 2^{15}\right)$;
	\item half the width of epsilon-insensitive band $\epsilon\in\left(0.1:0.1:11\right)$;
	\item Kernel Function \texttt{'rbf'}.  
\end{itemize}
\indent For GPR methodology, the considered hyperparametes are:
\begin{itemize}
	\item initial value for the noise standard deviation of the Gaussian process model\\ \begin{center}$\sigma\in\left(1e-2std(target\_train), \frac{std(target\_train)}{\sqrt{2}}\right)$;\end{center}
	\item Kernel (Covariance) Function$\in(\texttt{'squaredexponential'}, \texttt{'matern32'},\texttt{'matern52'})$\footnotemark[2]. 
\end{itemize}
\footnotetext[2]{\texttt{Squared Exponential Kernel} is one of the most commonly used covariance functions; it is defined as $k\left(x_l,x_j|\theta\right)=\sigma^2_f\exp\left[-\frac{1}{2}\frac{(x_l-x_j)^T (x_l-x_j)}{\sigma^2_l}\right]$. \texttt{Matern32} is a covariance function defined as $k\left(x_l,x_j|\theta\right)=\sigma^2_f\left(1+\frac{\sqrt{3r}}{\sigma_l}\right)\exp\left(-\frac{\sqrt{3r}}{\sigma_l}\right)$. \texttt{Matern52} is defined as $k\left(x_l,x_j|\theta\right)=\sigma^2_f\left(1+\frac{\sqrt{5r}}{\sigma_l}+\frac{5r^2}{\sigma_l}\right)\exp\left(-\frac{\sqrt{5r}}{\sigma_l}\right)$, where $r=\sqrt{(x_l-x_j)^T(x_l-x_j)}$ is the Euclidean distance between $x_l$ and $x_j$.} 
For RC, the parameters considered are:
\begin{itemize}
	\item $\rho\in\left(0.1:0.1:1\right)$,
	\item $IS\in\left(0.1:0.1:0.9\right)$,
	\item $RU\in\left[10, 20, 30, 50, 100, 150, 200, 250\right]$.
\end{itemize}

\section{Results}\label{sec:risultati}
\begin{figure}[H]
  	\centering
  	\includegraphics[height=250pt,width=400pt]{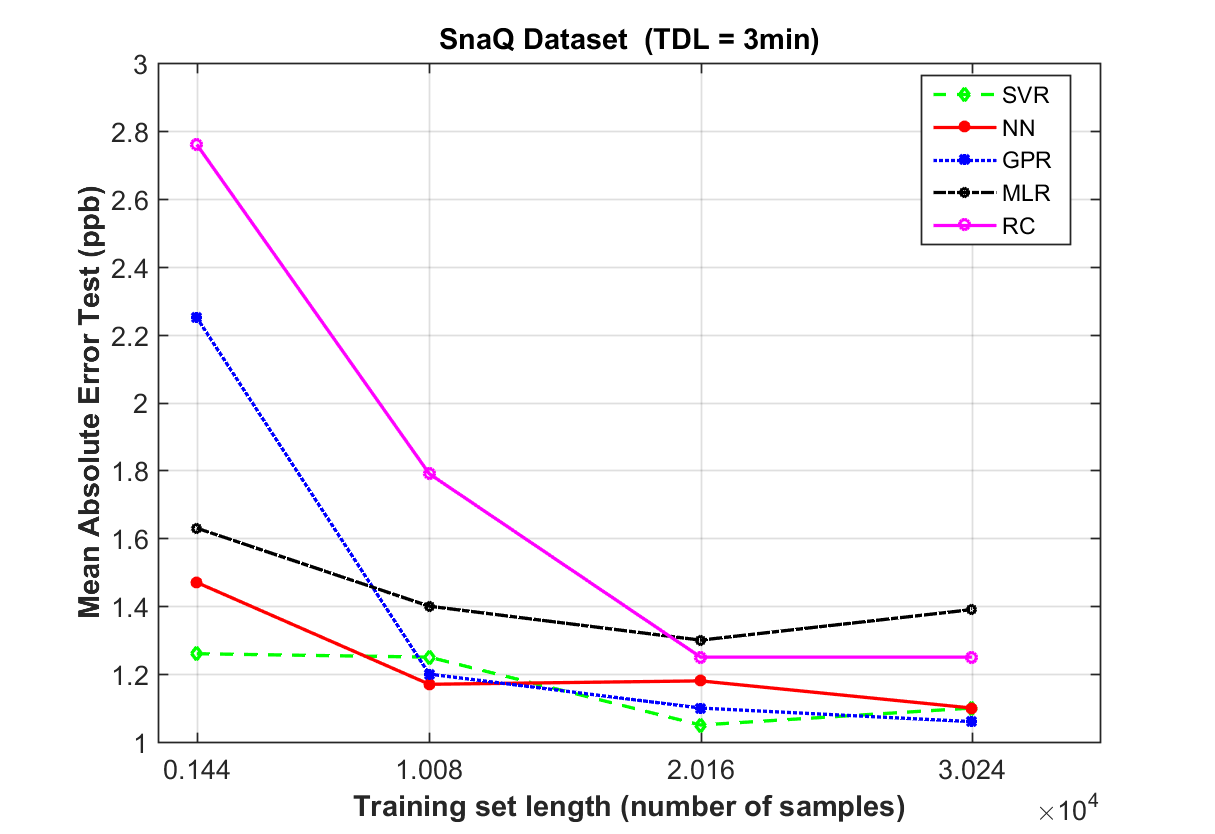}
  	\caption{MAE depending on Training Set length for $TDL=3 min$.}
  	\label{tslSnaq}
  \end{figure}

\begin{figure}[H]
  	\centering
  	\includegraphics[height=250pt,width=400pt]{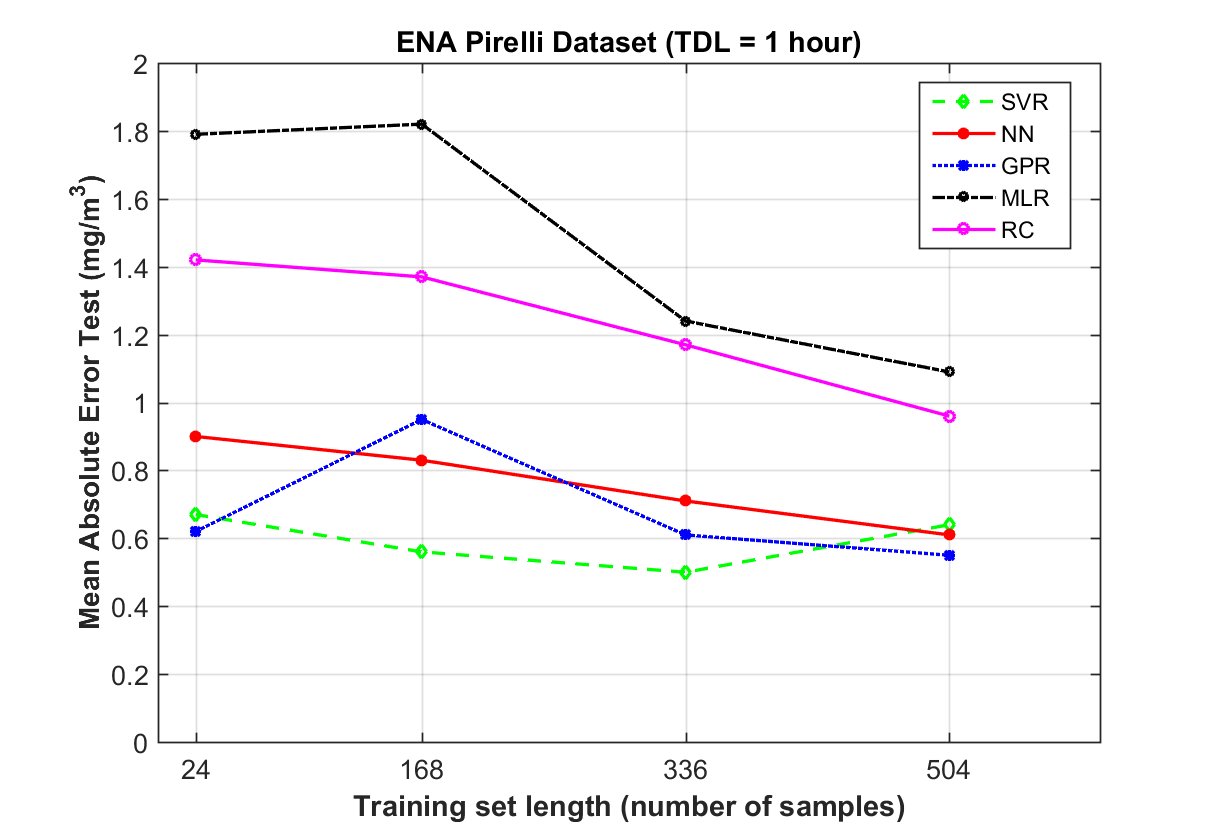}
  	\caption{MAE depending on Training Set length for $TDL=1 h$.}
  	\label{tslRosselli}
  \end{figure}

\begin{figure}[H]
  	\centering
  	\includegraphics[height=250pt,width=400pt]{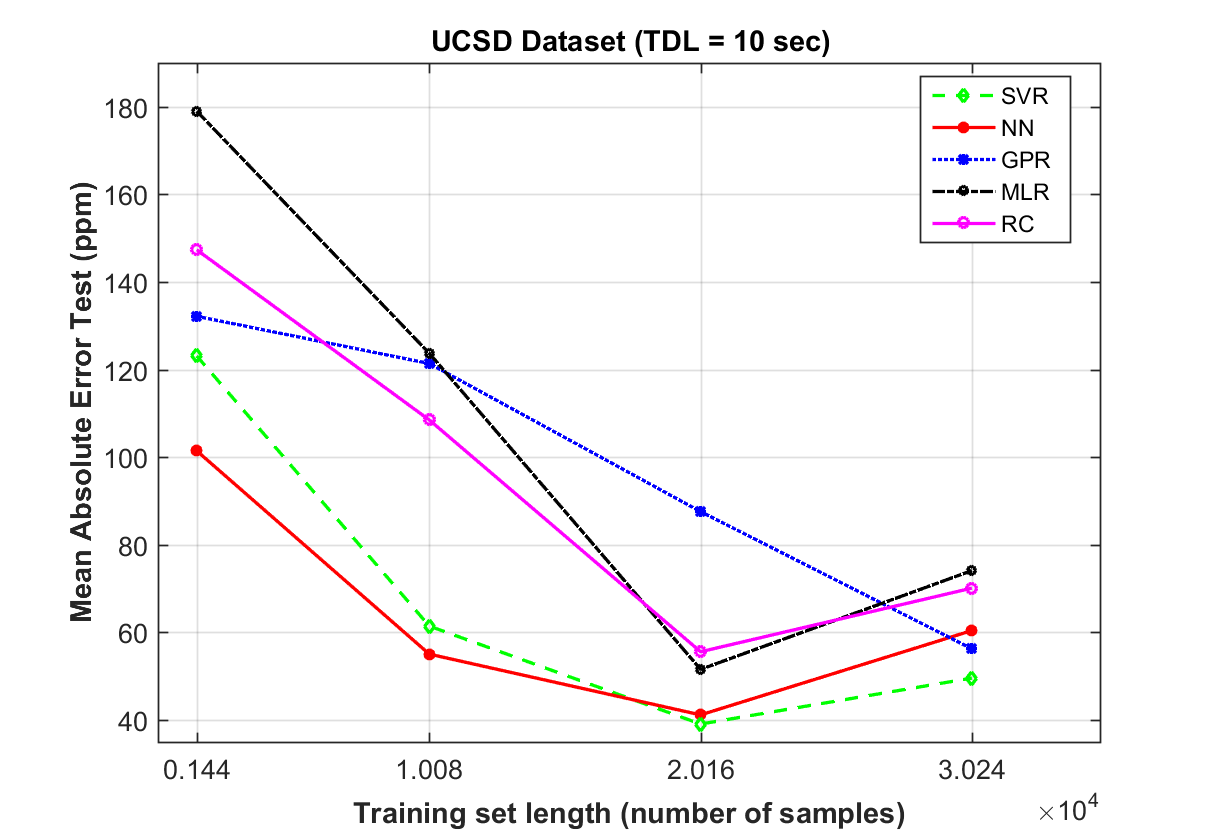}
  	\caption{MAE depending on Training Set length for $TDL=10 sec$.}
  	\label{tslFonollosa}
  \end{figure}

\begin{figure}[H]
  	\centering
  	\includegraphics[height=250pt,width=400pt]{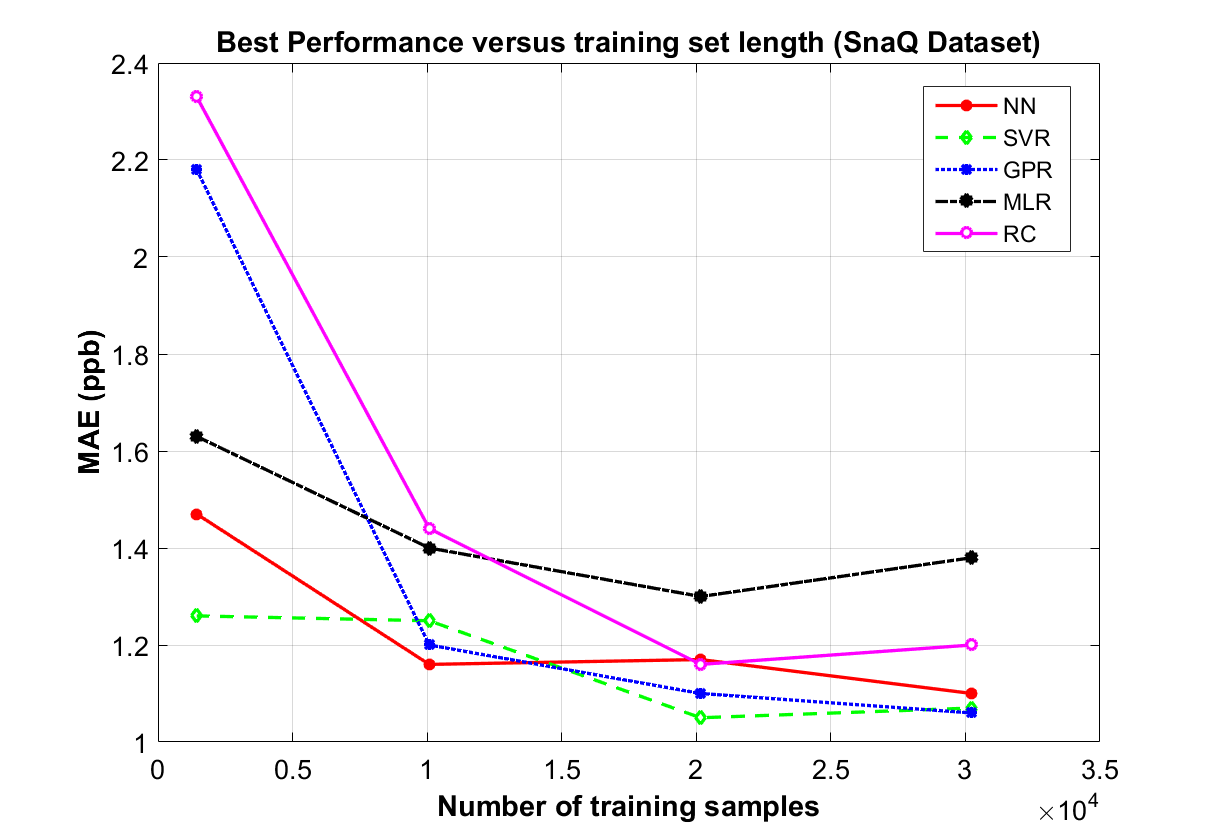}
  	\caption{Test set computed MAE depending on Training Set length. Exact trends are reported for each machine learning methodology at best performing hyper-parameters and TDL values for the SNAQ dataset.}
  	\label{BPversTSLSnaQ}
  \end{figure}

\begin{figure}[H]
  	\centering
  	\includegraphics[height=250pt,width=400pt]{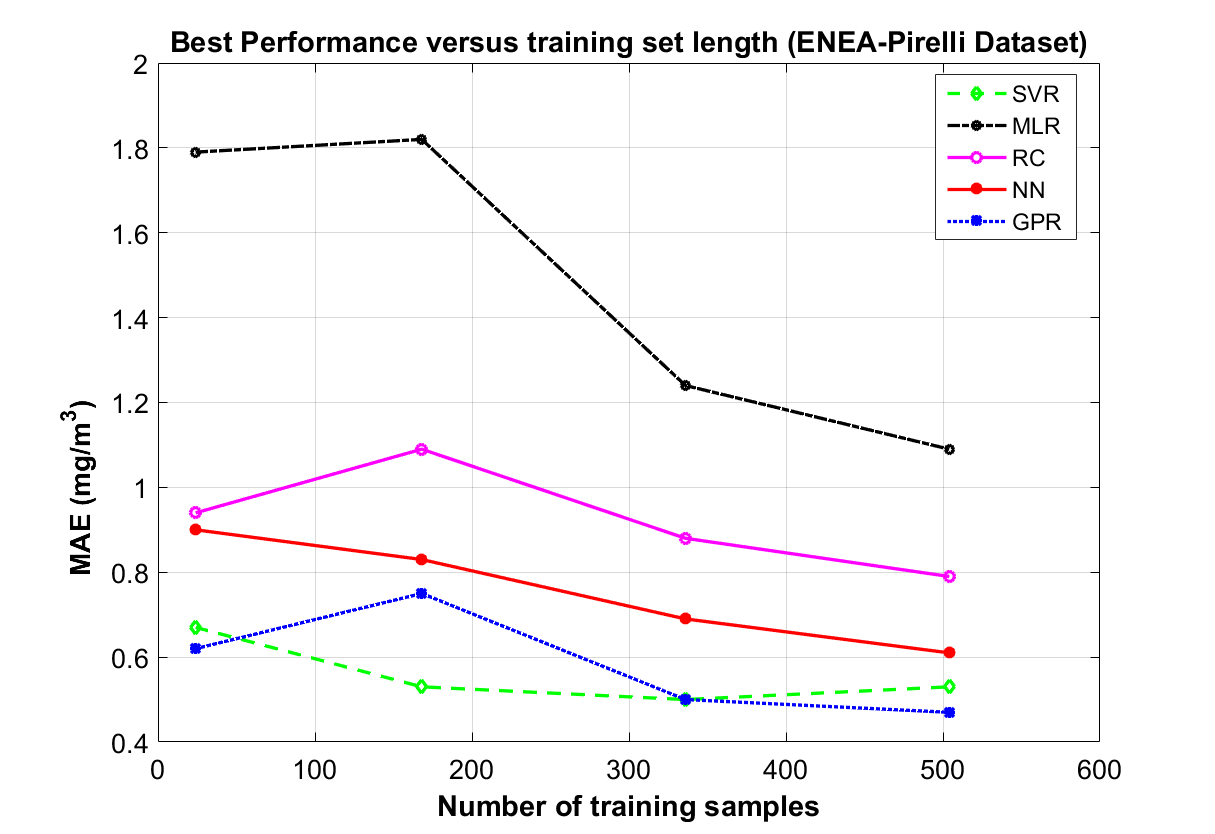}
  	\caption{Test set computed MAE depending on Training Set length. Exact trends are reported for each machine learning methodology at best performing hyper-parameters and TDL values for the Enea-Pirelli dataset.}
  	\label{BPversTSLPirelli}
  \end{figure}

\begin{figure}[H]
  	\centering
  	\includegraphics[height=250pt,width=400pt]{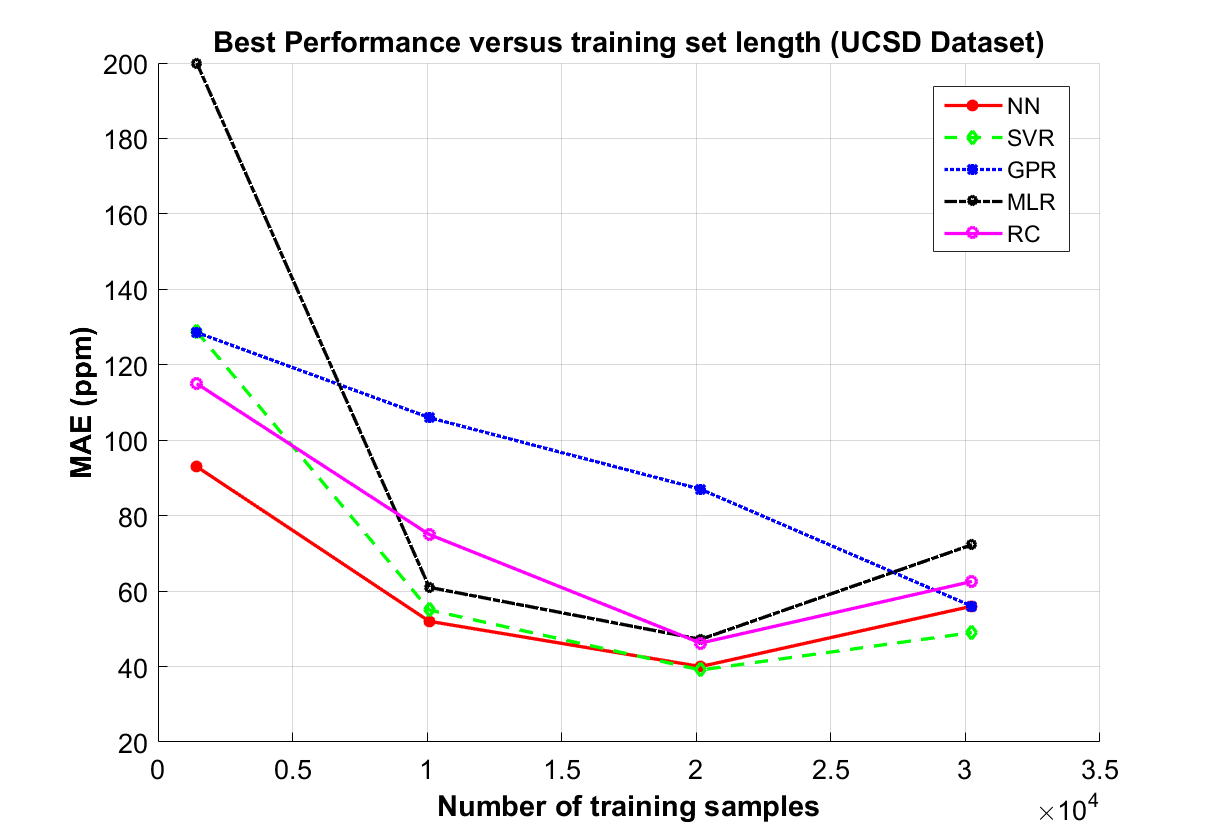}
  	\caption{Test set computed MAE depending on Training Set length. Exact trends are reported for each machine learning methodology at best performing hyper-parameters and TDL values for the UCSD dataset.}
  	\label{BPversTSLSnaDiego}
  \end{figure}

\begin{figure}[H]
	\centering
	\includegraphics[height=250pt,width=420pt]{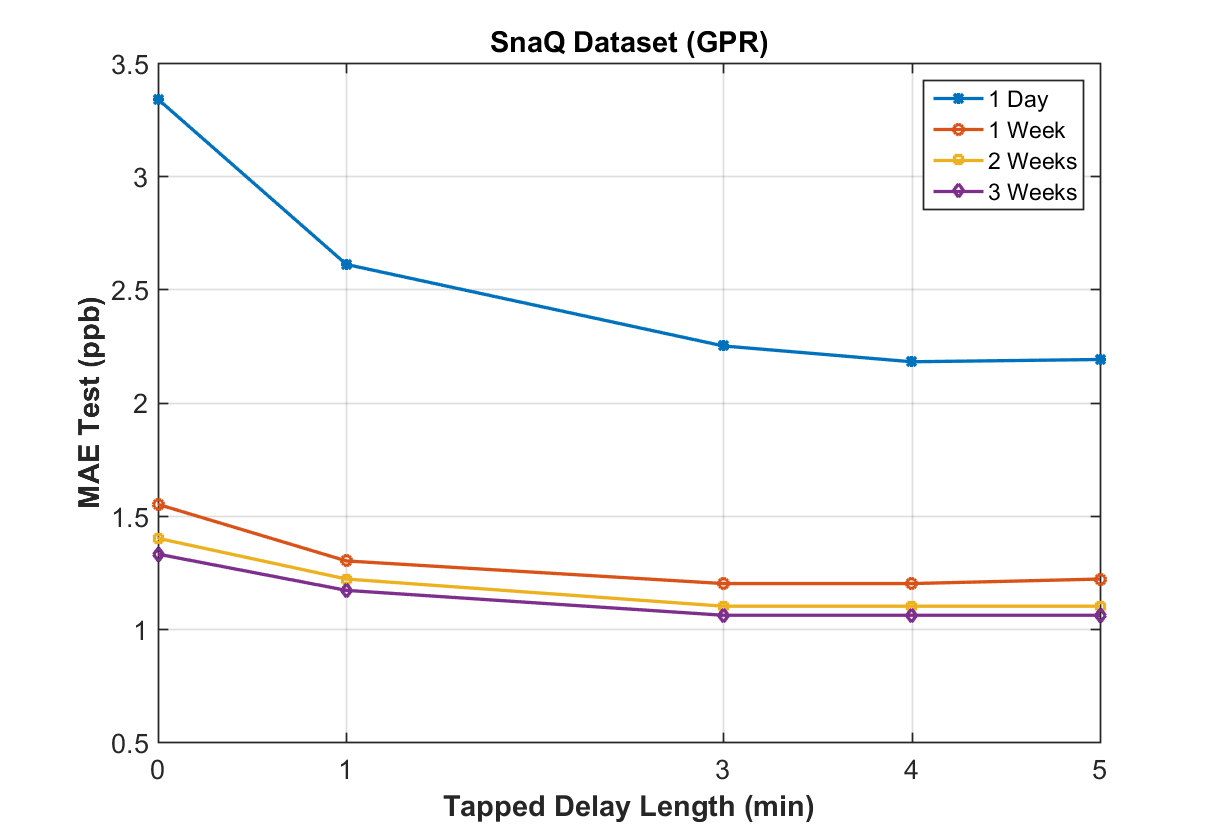}
	\caption{MAE obtained on test sets versus TD length for the GPR architectures for different training set length in the SNAQ dataset. Dynamic architectures show better performances with growing length of the observation window irrespective of the training set length employed (1 day, 1 week, 2 weeks or 3 weeks).} 
	\label{labsnaq}
\end{figure}

\begin{figure}[H]
	\centering
	\includegraphics[height=250pt,width=420pt]{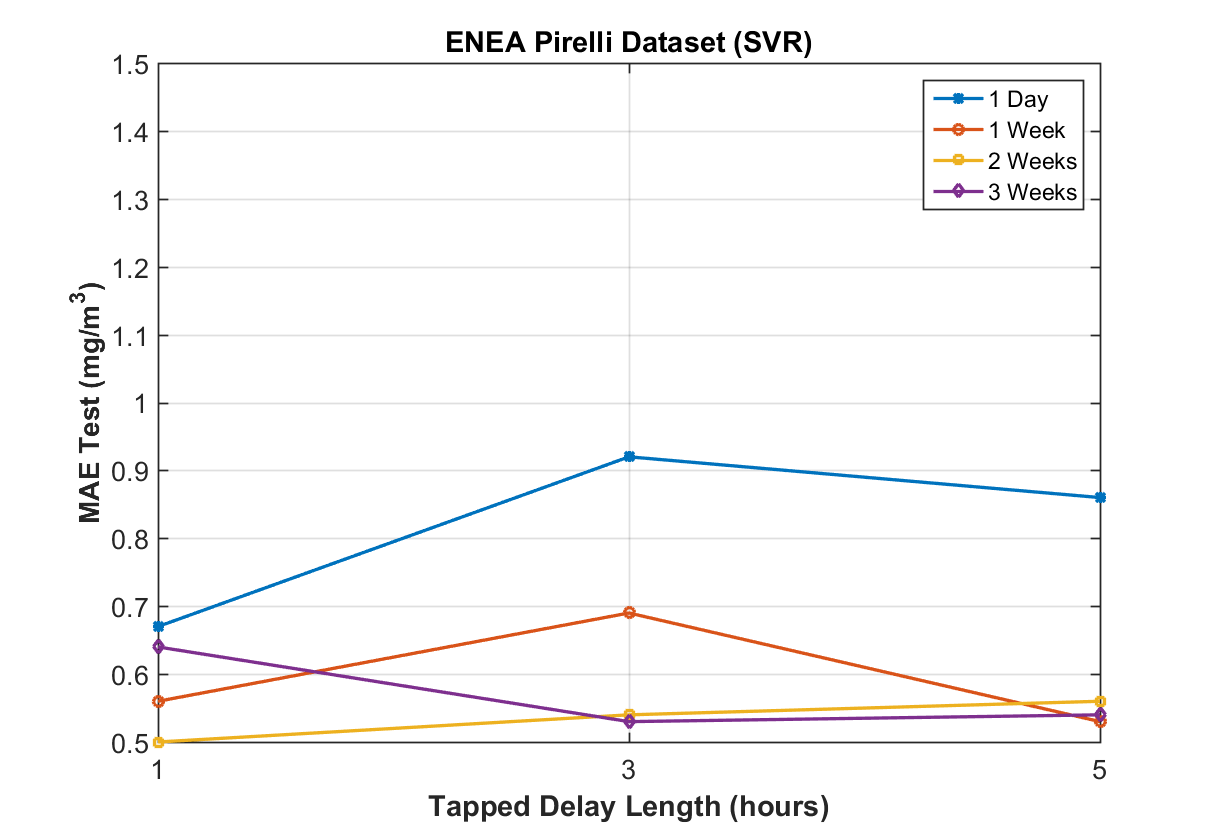}
	\caption{MAE obtained on test sets versus TD length for the SVR architectures for different training set length in the ENEA-Pirelli dataset. The dynamic architectures has no obvious advantage.}
	\label{labrosselli}
\end{figure}

\begin{figure}[H]
	\centering
	\includegraphics[height=250pt,width=420pt]{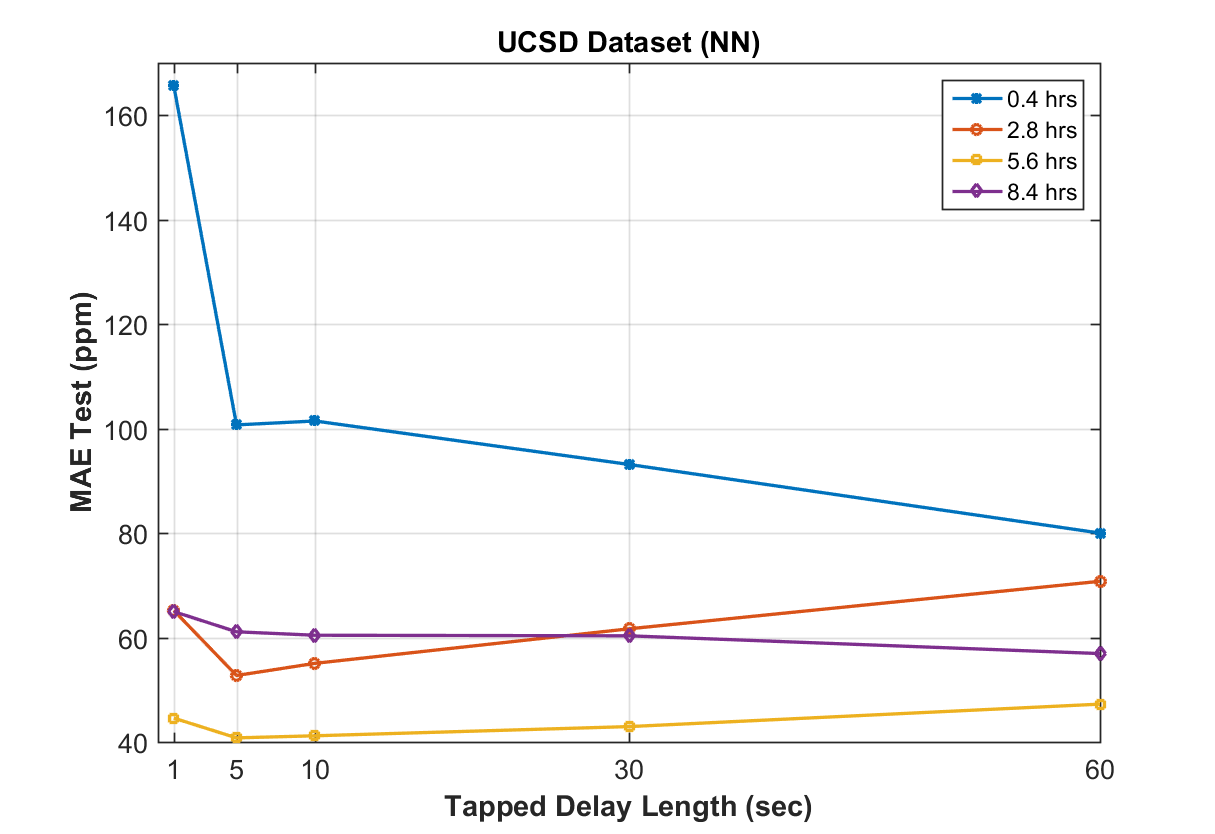}
	\caption{MAE obtained on test sets versus TD lengths for the neural networks architectures for different training set lengths in the UCSD dataset. Dynamic architectures show better performances irrespective of the employed training set length  (0.4hrs, 2.8hrs, 5.6hrs, 8.4hrs), starting from a 5 sec long observation window.}
	\label{labfonollosa}
\end{figure}

As mentioned above, all the methodologies taken into account in this work have been tested with different training set lengths (TSL) and different observation windows lengths (TDL). The comparison results are reported in Table \ref{tab:megatab}. 
In particular, there we show MAE and STD as performance indicators, for all the best performing models in the different scenarios (TSL,TDL). For the first dataset, the target gas was $NO_2$ (ranging from 0.30 to 48.50 ppb) and MAE estimations are expressed in $ppb$. For the second dataset, the results obtained for CO target gas estimation are in $mg/m^3$ while for the third dataset, CO was expressed in $ppm$. 
Graphically, Table \ref{tab:megatab} is divided into three main sections, each one reporting results for the three different datasets. In each section different subsection reports the results of the different ML techniques. Different rows indicate different training, validation and test set lengths. As mentioned above, we also compared the static architectures with the corresponding dynamic version, using different time series length (columns in \ref{tab:megatab}). Results from the static versions are reported in the leftmost sub column of the third main column (Tapped Delay Length) for each dataset and for all the proposed techniques except for reservoir computing. For its structure, RC is inherently a dynamic technique   and it is not coupled to a tapped delay with a specific length, instead its dynamic behaviour is controlled by the reservoir dimension. For this reason its results are reported in a single coloumn. At a first glance, we observe that dynamic versions of the machine learning architectures outperform, almost every time, their static counterparts. In fact, for the first and third dataset, regardless of the training set length and the applied ML methodology, the use of an observation window always gives best results. The absence of a consistent performance improvement in the second dataset is due to the sampling methodology (hourly averages) that average out any sensor related dynamic information from the dataset. By looking at performance dependence from the length of the tapped delay, in the first and third dataset, we can find a very interesting consistency in the minimal length required in order to boost static performance. Specifically, for the first dataset, we observe that best performances, regardless of the length of the training set, are obtained, for each methodology, at a minimum observation window length of 3 (SVR, GPR, MLR) or 4 minutes (NN). For the third dataset a similar behaviour is observed and most of the best performances are obtained for each training set length with an observation window of 30 seconds considering all the ML methodologies. This is graphically expressed by Fig. \ref{labsnaq}, \ref{labrosselli} and \ref{labfonollosa} that, respectively, shows the MAE performance versus TDL relationship for GPR (dataset 1), SVR (dataset 2) and NN (dataset 3), at each training set length. The pictures clearly highligh the MAE reductions obtainable by using dynamic architectures. These findings suggest that the dynamic architectures are truly capable to grasp and embed a useful knowledge on sensors dynamic behaviour provided the adoption of a sufficiently long observation window. Of course, curse of dimensionality issues may apply when a too large observation window is adopted. Further analyses are needed to correlate the observed results on optimal length with the time constants or, better, the T90 parameter of the slowest sensor in the sensor array. This could lead to a more concise feature set or, in other words, smaller input dimensionality and consequently a more concise representation. Summarizing these results, it seems clear that dynamic architectures  prove to be the best approach whenever prompt responsiveness is needed. 
The results obtained in Table \ref{tab:megatab} highlight (in bold) the best methodology for each dataset at each (TDL, TSL) setting. For SNAQ dataset SVR with training set length equal to 3 weeks and $TDL=3min$ gave the best overall results, i. e. $MAE=1.05$ $ppb$ compared to $MAE=1.10$ $ppb$ obtained with NN with three weeks long training set. Dividing this value by the experimental range of the $NO_2$ target gas, SVR achieved a relative MAE of $2.2\%$. For ENEA Pirelli dataset, the best results are obtained using GPR ($MAE=0.47$ $mg/m^3$ at three weeks training set accounting for a relative MAE of $3.9\%$). For the third dataset (UCSD) the best technique is once again SVR, with $MAE=39.10$ $ppm$ at 5.6 hr long training set scilicet a relative MAE of $7.3\%$.
Considering the relative performances of the individual techniques, we observe that in the first dataset, SVR architectures gave the best result for most training set length (3 out of 4 times). Furthermore considering each (TSL, TDL) setting separately, similar deduction can be made with SVR outperforming 12 times out of 20. However performance obtained by NN and GPR are relatively similar (see Table \ref{tab:megatab}). For the second dataset, we observe that SVR and GPR have very similar performance obtaining the best scores respectively 5 and 7 times out of 12. In the third dataset, both NN and SVR gave similar resulting to be sthe best performing techniques respectively 8 and 11 times out of 20, with GPR yielding the poorest performance. Thus, SVR seems to provide consistently the best performances. GPRs and, closely, NNs proved to be the next more reliable in terms of performance showing mixed behaviour when operating with a limited knowledge. Surprisingly, RC models, even if outperforming MLR, seems less reliable on average when compared to more classic approaches.  
\newline\indent In this case, we would also consider the significant variance due to the random ''training'' of the reservoir connections and the limited time that RC networks need to be designed and trained. Actually, the adoption of an RC technique simplifies modelling in that no selection of optimal observation window is needed, thereby exploiting the inherently dynamic wiring of the reservoir neurons. Moreover, redundancy in the reservoir dimension allows for enhanced flexibility. It allows, specifically,  the same architecture to operate efficiently with different scenarios showing different time constants or with a single scenario where different dynamics are simultaneously involved. In RC, adaptation take place in the training phase involving the final multi-linear stage by selecting the weight coefficients associated to the different components of the RC time expansion stage. In fact, looking at minimum MAE estimates from RC approach at each TSL, it appears that RC come closer to the performance obtained by NN, SVR and GPR at the longest TDL, outperforming them at least once (third dataset, 5.6 hrs long training set). 
\newline \indent Except under the conditions TDL=30sec, TSl=2.8 hrs for the third dataset, MLR showed the poorest relative performance. \newline \indent In order to consider the viability of implementing the above prediction algorithms on-board the devices considered, we have to evaluate our results from the computational and storage complexity point of view. From the previous sections, we know that the main challenges for the methodologies evaluated in this work are input dimensionality and training set length. Ideally we would like to obtain sufficient performances with limited input dimensionality (i.e. small number of sensing units and small window of observation) and limited training set length. Figures \ref{tslSnaq}, \ref{tslRosselli} and \ref{tslFonollosa} show the performance relationship with the TSL at the identified optimal TDL for the three datasets. Figures \ref{BPversTSLSnaQ}, \ref{BPversTSLPirelli} and \ref{BPversTSLSnaDiego} show the best performance obtained by the machine learning techniques for all the TDL with respect to the TSL for the three studied datasets.
  
\begin{figure}[H]
	\centering
	\includegraphics[height=280pt,width=400pt]{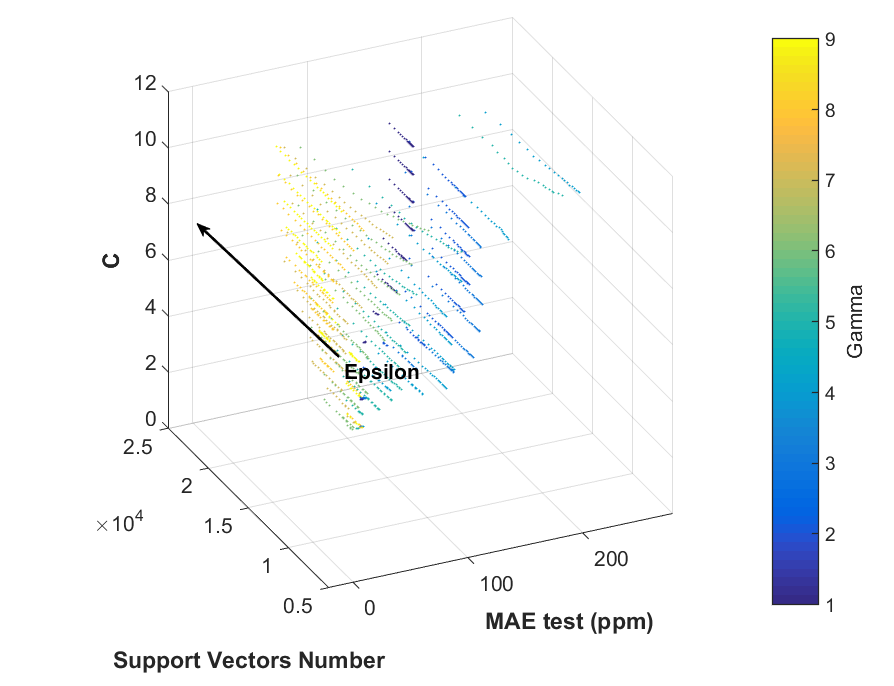}
	\caption{3-D scatterplot of Performance/Hyperparameters ($C, \gamma,\epsilon$)/Support Vectors number relationship for SVRs architectures in the UCSD dataset at best performing (TDL, TSL) setting. Note that a significant reduction of the number of needed support vectors can be obtained by fine tuning the C parameter value. }
	\label{grafico3d}
\end{figure}

\begin{figure}[H]
	\centering
	\includegraphics[height=280pt,width=400pt]{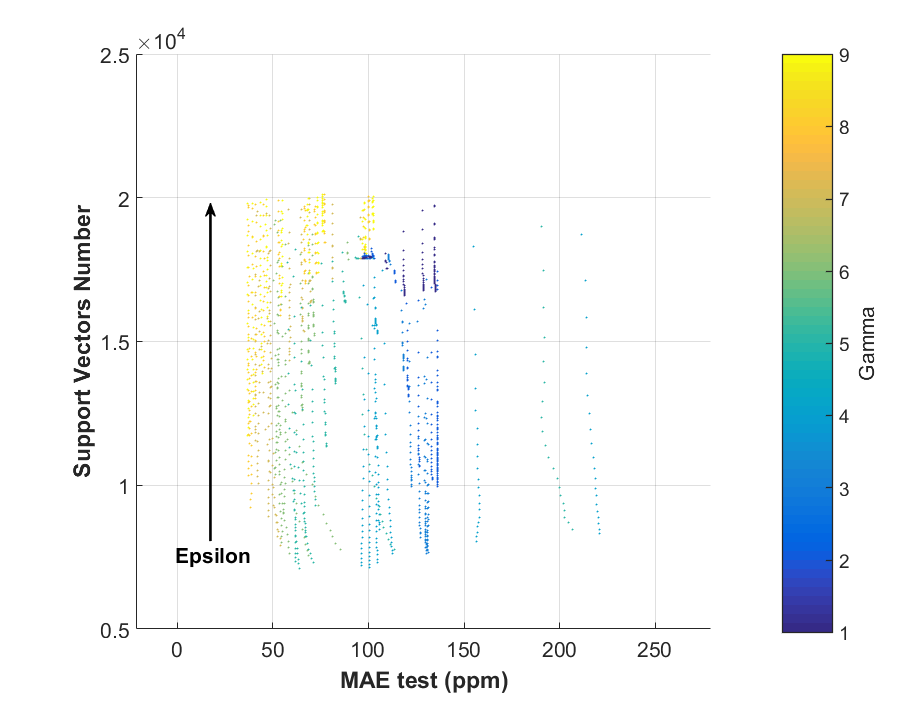}
	\caption{MAE/SV 2-D projection of the previous 3D scatterplot.}
	\label{grafico2d}
\end{figure}

Apart from confirming the relative performance among the different techniques, it is quite evident that, as already reported in literature, larger training sets generate better performance than very small one, the latter being unable to represent the real span of the multivariate manifold of sensor responses. However, far from being linear, this relationship usually shows that after a specific threshold, no significant improvement is attained by using larger training set. It is therefore important to identify this threshold value, in order to optimize the length of the training set and ultimately the cost of the calibration. 
Table \ref{table:resParam} shows the best performing hyper-parameters set at optimal TDL and TSL. Comparing the architectures obtaining the best performances for the different ML methodologies, our results show that SVRs hold a significant space complexity. Considering 3 weeks long training set of the SNAQ data, to achieve its best performance will require the storage of $10k\times d$ floating point precision values, where $d$ is the dimension of the input space. Since the number of support vectors both affects computational and storage complexity of the model, this can discourage their use as an on-board computational intelligence component for concentration estimation. It is worth noting that a high number of support vectors or, better still, high SVs to total number of training samples ratio, are quite common when operating performance-optimal parameter selection schemes. This is generally due to the peculiar choice of parameters. Considering figures \ref{grafico3d} and \ref{grafico2d}, depicting UCSD dataset (third dataset) results, we note how a significant reduction of the number of SV (up to one half) can be achieved at a small performance cost by tuning the C value. Nonetheless, the number of SV remains very high, suggesting the complex nonlinear nature of the calibration problem. Conversely, its corresponding optimal shallow neural network, being only bound to the number of hidden neurons and inputs dimensionality, needs, in our settings, a fraction of the number of the parameters (weights) to be stored. This confirms the Tapped Delay Neural Networks (TDNNs) as a very efficient methodology from the point of view of the learnt knowledge representation. Meanwhile, RC networks are bound by the number of hidden interconnected neurons in the reservoir. Together with input and output dimension, it generates the dimension of the input weights, output weights and internal interconnection weights matrices. In our scenario, they do not appear as a competitive choice. In fact, their optimal hyperparameters selection  needs considerably more storage space and computational capabilities than their optimal NN counterparts. However, they have shown better storage efficiency than optimal SVR architectures in most of the cases considered in this work. Gaussian processes, like all other \textit{lazy learning} machine learning tools, pose significant computational and storage issues when dealing with large datasets. Unlike SVRs, they need to store all the training set samples and use them for prediction computations, at least in their original formulation. Under this condition, the model can be used only when the number of available training samples is not high ($<<1000$ samples). In our model setup, specifically in the ENEA Pirelli dataset, GP use when using only 24 training samples, can be favourable also from the performance point of view. In some cases, the loss of performance can be justified by the limited computational and storage needs of a less than optimal methodology. Specifically, capability limits of the target microcontroller node or scalability issues when cloud computing is concerned, can be particularly relevant and may lead to such a choice. Most of the time, in arriving at the best option, engineers will have to take into account the trade-off between performance and complexity. In our view, considering the overall calibration scenario characterized by low dimensional feature sets and training set that ranges from hundreds to thousands samples, shallow neural networks appears to be the best option for on board integration. Actually, if we analyze the worst case for NNs, i.e. when $TDL=60$ (at $TSL=5.6hrs$) for the UCSD dataset, we observe that the difference between NN and SVR is negligible taking into account the computational complexity. In fact, the best estimation obtained with SVR involves the use of $10072$ support vectors, while NN required the storage of $2883$ weights. 
 \newpage
\begin{center}
	\tiny\begin{longtable}[H]{|c|p{10mm}|p{25mm}|c|c|c|c|c|}
		\hline
		\label{tab:megatab}
		\multirow{15}*{\rotatebox{90}{\textbf{SnaQ dataset} \qquad\qquad\qquad\qquad\qquad\qquad}}	
		& \centering\textbf{ ML technique} & \centering \textbf{(Train - Val - Test) Partition} & \multicolumn{5}{c|}{\textbf{Mean Absolute Error Test (STD)}}\\ 
		\hline
		& & (Number of samples)& \multicolumn{5}{c|}{\textbf{ Tapped Delay Length (min)}}	\\
		\hline
		\rowcolor{gray!30}
		&  &  & {0.33} & {1} & {3} & {4} & {5} \\
		\hline
		& \multirow{5}{*}{NN}
		&   1440 - 10080 - 40285 & 1.62{ (0.08)} & 1.57{ (0.13)} & 1.47{ (0.11)} & 1.54{ (0.14)} & 1.53{ (0.14)} \\ \cline{3-8}
		& & 10080 - 10080 - 31645 & \textbf{1.26{ (0.05)} }& \textbf{1.19{ (0.06)}} & \textbf{1.17{ (0.04)}} & \textbf{1.16{ (0.04)}} & 1.16{ (0.04)} \\ \cline{3-8}
		& & 20160 - 10080 - 21565 & 1.39{ (0.04)} & 1.32{ (0.03)} & 1.18{ (0.03)} & 1.17{ (0.03)} & 1.18{ (0.02)} \\ \cline{3-8}
		& & 30240 - 10080 - 11485 & 1.38{ (0.04)} & 1.24{ (0.03)} & 1.10{ (0.02)} & 1.10{ (0.02)} & 1.10{ (0.03)}\\ \cline{2-8}
		& & & & & & &\\ \cline{3-8}
		& \multirow{5}{*}{SVR} 
		&  \centering   1440 - 10080 - 40285 & \textbf{1.41} &\textbf{1.34} & \textbf{1.26} & \textbf{1.29} & \textbf{1.28}\\ \cline{3-8}
		& & \centering 10080 - 10080 - 31645 & 1.51 & 1.30& 1.25& 1.25& \textbf{1.16} \\ \cline{3-8}
		& & \centering 20160 - 10080 - 21565 & \textbf{1.31} &\textbf{1.15} &\textbf{1.05} &\textbf{1.05} &\textbf{1.05} \\ \cline{3-8}
		& & \centering 30240 - 10080 - 11485 & \textbf{1.32} & 1.18 & 1.10 &1.07 &1.07 \\ \cline{2-8}
		& & & & & & &\\	\cline{3-8}
		& \multirow{5}{*}{GPR} 
		&   \centering  1440 - 10080 - 40285 &  3.34  { (0) } & 2.61 {(0.05)}&2.25 {(0.17)}& 2.18 {(0.20)}& 2.19 {(0.25)}\\ \cline{3-8}
		& & \centering 10080 - 10080 - 31645 &1.55{ (0.02)} &1.30 {(0.02)} & 1.20{ (0.02)}& 1.20{ (0.01)}& 1.22{ (0.01)} \\ \cline{3-8}
		& & \centering 20160 - 10080 - 21565 & 1.40{ (0.01)}  & 1.22{ (0.05)}& 1.10{ (0.03)} & 1.10 {(0.03)} &1.10{ (0.03)} \\ \cline{3-8}
		& & \centering 30240 - 10080 - 11485 &1.33 {(0.01)} & \textbf{ 1.17 {(0.004)}}& \textbf{  1.06 {(0.002)}} &   \textbf{  1.06 {(0.003)}} &\textbf{  1.06 {(0.002)}}\\ \cline{2-8}
		& & & & & & &\\	\cline{3-8}
		& \multirow{5}{*}{MLR}
		&  \centering   1440 - 10080 - 40285 &  1.81 &1.66 &  1.63& 1.65&1.67 \\ \cline{3-8}
		& &\centering 10080 - 10080 - 31645 & 1.62 &1.47 &1.40 &1.40 &1.40 \\ \cline{3-8}
		& &\centering 20160 - 10080 - 21565 & 1.55 &1.40 &1.30 &1.30 &1.30 \\ \cline{3-8}
		& &\centering 30240 - 10080 - 11485 & 1.58 &1.48 &1.39 &1.38 &1.38 \\ \cline{2-8}
		& & & & & & &\\	\cline{3-8}
		& \multirow{5}{*}{RC} 
		&  \centering 1440 - 10080 - 40285  & \multicolumn{5}{c|}{3.02{ (0.31)}/2.33}	\\ \cline{3-8}
		& & \centering 10080 - 10080 - 31645 & \multicolumn{5}{c|}{2.74{ (0.95)}/1.44}	\\ \cline{3-8}
		& & \centering 20160 - 10080 - 21565 & \multicolumn{5}{c|}{1.55{ (0.76)}/1.16}	\\ \cline{3-8}
		& & \centering 30240 - 10080 - 11485 & \multicolumn{5}{c|}{1.25{ (0.03)}/1.20}	\\ \cline{1-8}		
		
		\multirow{15}{*} {\rotatebox{90}{\textbf{ENEA Pirelli dataset \qquad\qquad\qquad\qquad\quad}}} 
		&  & (Number of samples)& \multicolumn{5}{c|}{\textbf{Tapped Delay Length (hours)}} \\ 
		\hline
		\rowcolor{gray!30}
		&  &  & {} & {1} & {3} & {5} & {} \\
		\hline
		& \multirow{5}{*}{NN} 
		& \centering 24 - 168 - 7482 & & 0.90{ (0.10)} &0.96{ (0.43)} & 1.22{ (0.37)}&  \\ \cline{3-8}
		& & \centering 168 - 168 - 7338 & & 0.83{ (0.11)}& 0.86{ (0.19)}& 0.79{ (0.14)} & \\ \cline{3-8}
		& & \centering 336 - 168 - 7170 & & 0.71{ (0.10)}& 0.79{ (0.13)}& 0.69{ (0.09)}& \\ \cline{3-8}
		& & \centering 504 - 168 - 7002 &  &0.61 {(0.06)}& 0.76{ (0.12)} & 0.71 {(0.12)}& \\ \cline{2-8}
		& & & & & & &\\	\cline{3-8}
		
		& \multirow{5}{*}{SVR} 
		&  \centering  24 - 168 - 7482 & & 0.67& 0.92& 1.17& \\ \cline{3-8}
		& & \centering 168 - 168 - 7338 & & \textbf{0.56}& \textbf{0.69}& \textbf{0.53}& \\ \cline{3-8}
		& & \centering 336 - 168 - 7170 & & \textbf{0.50} &0.54 &\textbf{0.56} & \\ \cline{3-8}
		& & \centering 504 - 168 - 7002 &  & 0.64& 0.53 & 0.54 & \\ \cline{2-8}
		& & & & & & &\\	\cline{3-8}
		
		& \multirow{5}{*}{GPR} 
		&  \centering  24 - 168 - 7482 &  & \textbf{0.62 {  (0) }}& \textbf{0.80 {(0.02)}}& \textbf{0.67 {(0.02)}}& \\ \cline{3-8}
		& & \centering 168 - 168 - 7338 & & 0.95{   (0) } & 0.75 {(0.09)} & 0.78 {(0.02)} & \\ \cline{3-8}
		& & \centering 336 - 168 - 7170 & & 0.61{(0.05)}& \textbf{0.50 { (0) }}& 0.75 {( 0)}& \\ \cline{3-8}
		& & \centering 504 - 168 - 7002 &  &\textbf{0.55{ (0)}} & \textbf{0.47{ (0) }}& \textbf{0.52{ (0) }}& \\ \cline{2-8}
		& & & & & & &\\	\cline{3-8}
		& \multirow{5}{*}{MLR}
		&  \centering  24 - 168 - 7482 & & 1.79 & - &  - & \\ \cline{3-8}
		& & \centering 168 - 168 - 7338 & & 1.82 & 1.97 & 1.70 &  \\ \cline{3-8}
		& & \centering 336 - 168 - 7170 & & 1.24 & 1.43 & 1.36 & \\ \cline{3-8}
		& & \centering 504 - 168 - 7002 & & 1.09 & 1.33 & 1.30 & \\ \cline{2-8}
		& & & & & & &\\	\cline{3-8}
		& \multirow{5}{*}{RC} 
		&  \centering 24 - 168 - 7482  & \multicolumn{5}{c|}{1.42 {(0.28)}/0.94}	\\ \cline{3-8}
		& & \centering 168 - 168 - 7338 & \multicolumn{5}{c|}{1.37 {(0.21)}/1.09}	\\ \cline{3-8}
		& & \centering 336 - 168 - 7170 & \multicolumn{5}{c|}{1.17 {(0.25)}/0.88}	\\ \cline{3-8}
		& & \centering 504 - 168 - 7002 & \multicolumn{5}{c|}{0.96 {(0.14)}/0.79}	\\ \cline{1-8}	
			
		\multirow{15}*{\rotatebox[origin=rB]{90}{\textbf{UCSD dataset \qquad\quad\qquad\qquad\qquad\qquad}}} 
		&  & (Number of samples)& \multicolumn{5}{c|}{\textbf{Tapped Delay Length (sec)}} \\ 
		\hline
		\rowcolor{gray!30}
		&  &  & {1} & {5} & {10} & {30} & {60} \\
		\hline
		
		& \multirow{5}{*}{NN} 
		&  \centering  1440 - 10080 - 30562 & 165.73 {(57.32)} & \textbf{100.71 {(45.68)}} & \textbf{101.46 {(24.96)}}& \textbf{93.14 {(20.76)}}& 144.13 {(66.68)} \\ \cline{3-8}
		& &\centering 10080 - 10080 - 21922 & 65.20 {(12.86)} & \textbf{52.73 {(6.50)}} & \textbf{55.06 {(7.10)}} & 61.67 {(10.15)} & \textbf{70.77 {(5.94)}}\\ \cline{3-8}
		& &\centering 20160 - 10080 - 11842 & 44.58 {(5.24)} & \textbf{40.79 {(1.41)} }& 41.19 {(3.01)} & 42.95 {(2.96)} & \textbf{47.25 {(4.16)}}\\ \cline{3-8}
		& &\centering 30240 - 10080 -  1762 & 64.98 {(2.68)} & 61.10 {(3.12)} & 60.44 {(3.29)} & 60.33 {(3.15)} & 56.93 {(2.81)}\\ \cline{2-8}
		& & & & & & &\\	\cline{3-8}
		& \multirow{5}{*}{SVR} 
		& \centering   1440 - 10080 - 30562 & \textbf{128.81} & 119.48 & 123.12 &135.40 &\textbf{125.95} \\ \cline{3-8}
		& &\centering 10080 - 10080 - 21922 & \textbf{55.76} &64.41 &61.46&74.37 &76.41 \\ \cline{3-8}
		& &\centering 20160 - 10080 - 11842 & \textbf{39.43} & 41.85 & \textbf{39.10} & \textbf{41.75}& 47.67\\ \cline{3-8}
		& &\centering 30240 - 10080 -  1762 & \textbf{57.97} & \textbf{53.19} & \textbf{49.59} & \textbf{53.52} & \textbf{54.79}  \\ \cline{2-8}
		& & & & & & &\\	\cline{3-8}
		& \multirow{5}{*}{GPR} 
		& \centering   1440 - 10080 - 30562 &  136.92 {(0.20)}& 133.27 {(0.07)}& 132.18{ (0.02)}& 130.76 {(0)}&128.57{ (0)} \\ \cline{3-8}
		& &\centering 10080 - 10080 - 21922 & 129.95 {(0)} & 125.87 {(0)}& 121.39 {(0)}& 106.55 {(0)}& 120.68{ (0)} \\ \cline{3-8}
		& &\centering 20160 - 10080 - 11842 & 89.92{(0)}& 91.69{ (0)}& 87.59{ (0)}& 91.23 {(0)}& 90.74 {(0)}\\ \cline{3-8}
		& &\centering 30240 - 10080 -  1762 &  67.38 { (0.17)}& 64.15 {(0.12)}& 56.27 {(0.09)}& 61.05 {(0)}& 59.04 {(0.03)} \\ \cline{2-8}
		& & & & & & &\\	\cline{3-8}
		& \multirow{5}{*}{MLR}
		& \centering   1440 - 10080 - 30562 & - & - & - & - & \\ \cline{3-8}
		& &\centering 10080 - 10080 - 21922 & 184.50 & 166.19 & 123.73 &\textbf{61.51} & 142.24 \\ \cline{3-8}
		& &\centering 20160 - 10080 - 11842 & 106.72 & 59.77 & 51.58 & 47.21 & 49.60 \\ \cline{3-8}
		& &\centering 30240 - 10080 -  1762 & 95.35 & 79.54 & 74.13 & 72.29 & 73.64 \\ \cline{2-8}
		& & & & & & &\\	\cline{3-8}
		& \multirow{5}{*}{RC} 
		&  \centering 1440 - 10080 - 30562  & \multicolumn{5}{c|}{147.35 {(39.76)}/115.10}	\\ \cline{3-8}
		& & \centering 10080 - 10080 - 21922 & \multicolumn{5}{c|}{108.51 {(15.15)}/75.06}	\\ \cline{3-8}
		& & \centering 20160 - 10080 - 11842 & \multicolumn{5}{c|}{55.60 {(6.38)}/46.22}	\\ \cline{3-8}
		& & \centering 30240 - 10080 -  1762 & \multicolumn{5}{c|}{70.13 {(3.64)}/62.55}	\\ \cline{1-8}	
		\caption{Listing of the results obtained by all the selected ML methodologies for all the considered datasets at different TDLs and four different training set lengths. Bold values indicates best performances for each (TDL,TSL) settings. "-" indicates out of scale values.}
	\end{longtable}
\end{center}
\begin{center}	
	\tiny{\begin{longtable}[t]{|c|c|c|c|}
			\hline
			\multicolumn{3}{|c|}{\textbf{Best Hyperparameters}}\\ 
			\hline
			&\centering\textbf{Best ML technique} &\textbf{Tapped Delay Length}\\ 
			\hline
			&   &       3 min  \\ \cline{3-4}
			\multirow{7}{*}{\rotatebox{90}{\textbf{SnaQ }\qquad\qquad\qquad\qquad}}
			& \multirow{4}{*}{SVR (30240 samples)}
			& \\
			& &     $\gamma=2^5$  \\
			&	&       $C=2^{13}$   \\
			&	&   $\epsilon=0.7$  \\
			&	&    $sv=13095$ \\
			& &   \\ \cline{2-4}
			
			& \multirow{2}{*}{NN (30240 samples)}
			&  \\
			& &    $Epochs=900$  \\
			&& $hn=15$  \\
			& &  \\ \cline{2-4}
			
			& \multirow{2}{*}{GPR (30240 samples)}
			& \\
			& &  $\sigma=0.2439$\\
			&	& $KF=squaredexp$ \\
			& &  \\ \cline{2-4}	
			
			& \multirow{3}{*}{RC (30240 samples)}
			&  \\
			& & $\rho=0.1$ \\
			& &  $input\_scaling=0.1$ \\
			& &   $output\_dim=250$ \\
			& &  \\ \cline{1-4}
			
			&   &  1 hour \\ \cline{3-4}
			\multirow{3}{*} {\rotatebox{90}{\textbf{ENEA Pirelli}\qquad\qquad\qquad\qquad}} 
			& \multirow{4}{*}{SVR (504 samples)}
			& \\
			& &     $\gamma=2^4$  \\
			&	&       $C=2^9$   \\
			&	&   $\epsilon=0.2$  \\
			&	&    $sv=356$ \\
			& &   \\ \cline{2-4}
			& \multirow{2}{*}{NN (504 samples)}
			& \\
			& &    $Epochs=400$  \\
			&	&   $hn=5$      \\
			& &  \\ \cline{2-4}			
			
			& \multirow{2}{*}{GPR (504 samples)}
			&  \\
			& &   $\sigma=0.0136$  \\
			&	&    $KF=Matern32$   \\
			& &  \\ \cline{2-4}
			
			& \multirow{3}{*}{RC (504 samples)}
			&  \\
			& &  $\rho=0.9$ \\
			& &  $input\_scaling=1e-9$    \\
			& &  $output\_dim=200$\\
			& &  \\ \cline{1-4}
			
			&   & 10 sec \\ \cline{3-4}
			\multirow{4}{*} {\rotatebox{90}{\textbf{UCSD}\qquad\qquad\qquad\qquad}}
			& \multirow{4}{*}{SVR (20160 samples)}
			&  \\
			& &    $\gamma=2^5$  \\
			&	&     $C=2^{15}$   \\     
			&	&   $\epsilon=1.3$ \\
			&	&    $sv=14447$  \\
			& &  \\ \cline{2-4}
			
			& \multirow{2}{*}{NN (20160 samples)}
			&  \\
			& &   $Epochs=600$ \\
			& &     $hn=3$  \\     
			& &  \\ \cline{2-4}
			& \multirow{2}{*}{GPR (20160 samples)}
			&  \\
			& &    $\sigma=91.6378$ \\
			&	  &  $KF=matern32$  \\
			& &  \\ \cline{2-4}	
			
			& \multirow{3}{*}{RC (20160 samples)}
			& \\
			& &  $\rho=0.3$ \\
			& &   $input\_scaling=0.01$     \\
			& &   $output\_dim=100$ \\
			& &   \\ \cline{1-4}
			\caption{Best parameters provided by the considered methodologies, for the three datasets at the best performing tapped delay length.}
			\label{table:resParam}
		\end{longtable}}
	\end{center}				
\newpage
 In Fig. \ref{grafico3d} and Fig. \ref{grafico2d} we show the trends of MAE while changing the SVR hyperparameters values. 
\subsection{Dynamic performances} 
\begin{figure}[H]
	\centering
	\includegraphics[height=250pt,width=420pt]{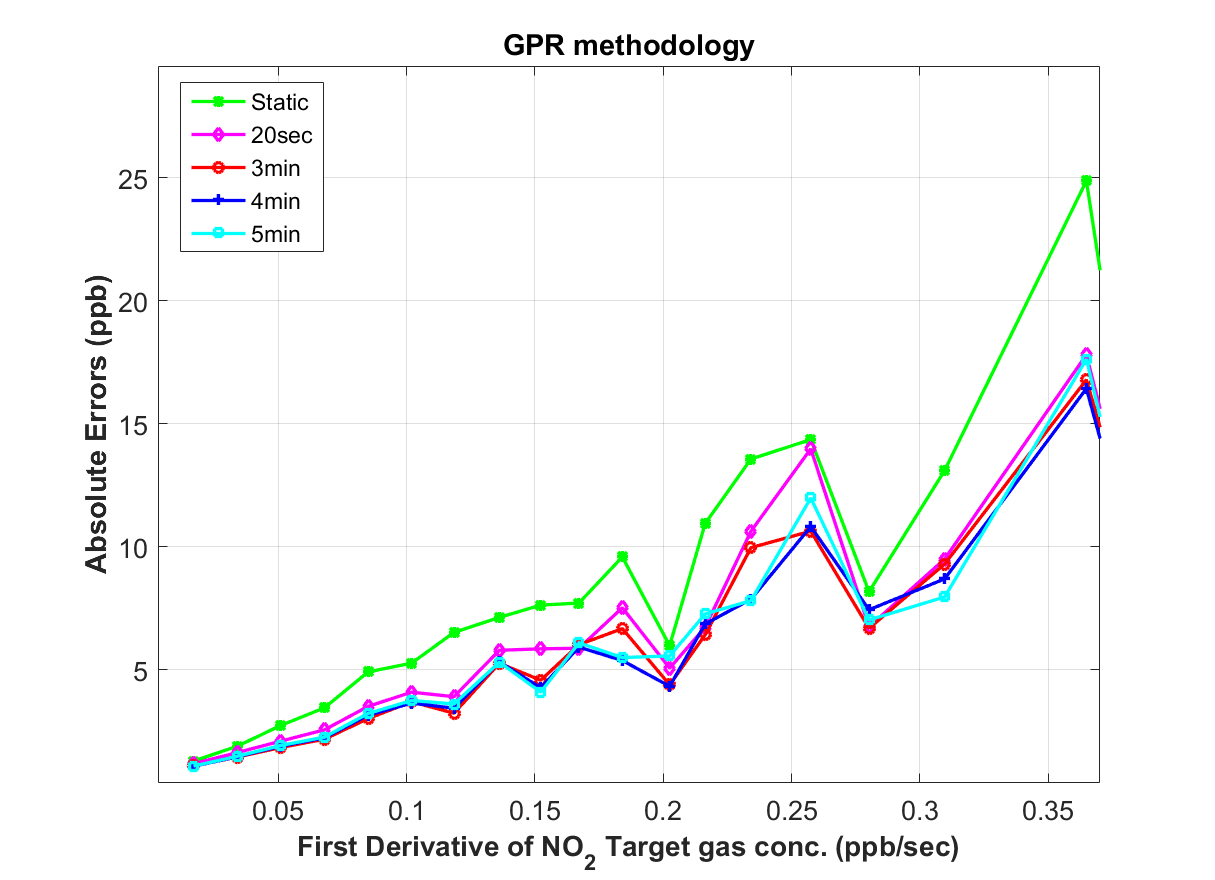}
	\caption{Comparison of average absolute error trends (SNAQ dataset) for increasing absolute derivative of the $NO_2$ target gas concentration for dynamic and static GPRs algorithms. We can clearly see the advantage of dynamic architectures over the static architecture.}
	\label{derivataSnaQ}
\end{figure}

\begin{figure}[H]
	\centering
	\includegraphics[height=250pt,width=420pt]{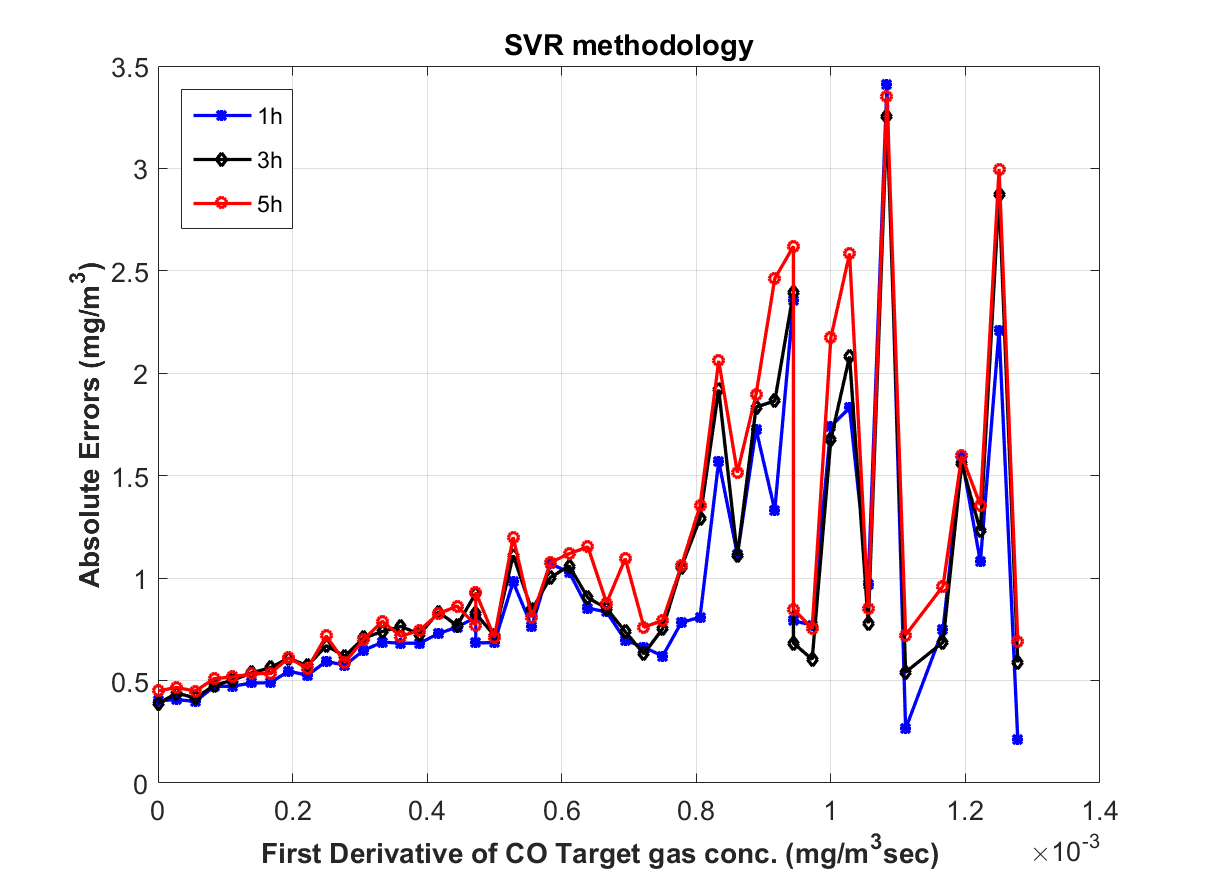}
	\caption{Comparison of average absolute error trends versus the absolute derivative of the CO target gas concentration (ENEA Pirelli Dataset) for dynamic (in black and red) and static (in blue) SVRs. Note the mean  MAE during rapid concentration changes are larger than those during slow concentration variations.. No advantage is obtained by using dynamic architectures.}
	\label{derivataRosselli}
\end{figure}

\begin{figure}[H]
	\centering
	\includegraphics[height=250pt,width=420pt]{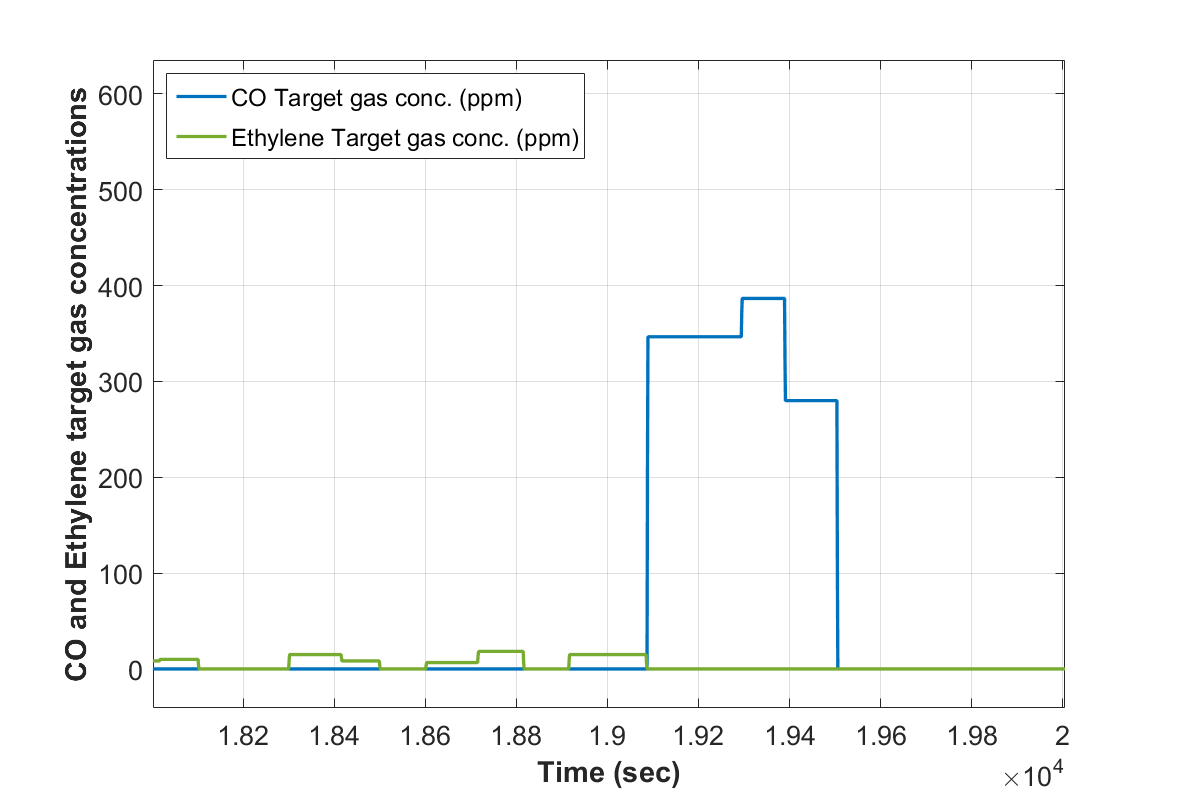}
	\caption{Sensors readings vs CO target gas concentration variation in the UCSD dataset (top panel). The response slow dynamic of the sensor array is clearly visible.}
	\label{sensoriFonollosa}
\end{figure}

\begin{figure}[H]
	\centering
	\includegraphics[height=250pt,width=420pt]{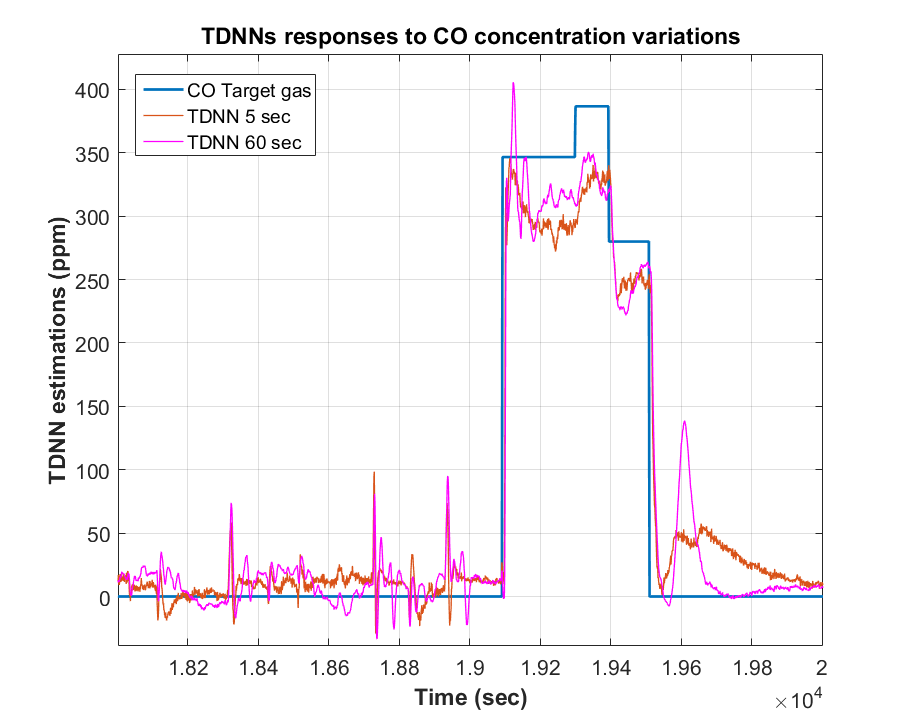}
	\caption{TDNNs responses to CO concentration variations in the UCSD dataset. The responses appear to resemble the forcing stimulus (blue) with good accuracy and fast rate. Note that over-elongations can be observed in the TDL=60s architecture (pink). In general, the estimations appear also to be affected by short term burst correlated to raw sensors  responses to interferents gases (see fig. \ref{sensoriFonollosa}). This may be a side effect of training the networks to respond to abrupt concentration transients.}
	\label{tdnn_fonollosa5e60}
\end{figure}

\begin{figure}[H]
	\centering
	\includegraphics[height=250pt,width=420pt]{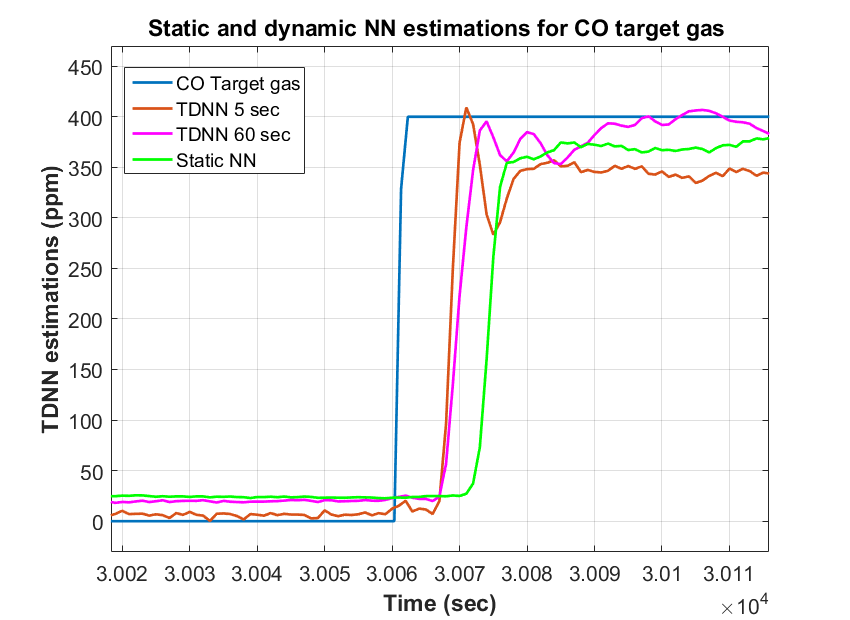}
	\caption{Comparison of concentration estimations using dynamic and static neural networks for CO species during an abrupt concentration change in the UCSD dataset. The two dynamic systems show faster response to the stimulus, the faster reaching 90 percent of the full response 6.6 seconds before the static one.}
	\label{comparisonNN}
\end{figure}

This section describes our results in view of the systems overall response to rapid changes of gas concentrations, to further analyze the outcome of the use of dynamic calibration algorithms. In particular, we are interested in following what happens at different rate of concentration change. To this effect, we have plotted the MAE performance indicator against the approximated value of first derivative of reference concentrations for the first two datasets ($NO_2$ species for SNAQ and $CO$ species for ENEA Pirelli). For the third dataset (UCSD), where reference concentrations set points changes abruptly, we have plotted the response to these  concentration changes allowing us to compare the static and dynamic approaches responses to rapid transients. Figures \ref{derivataSnaQ} and \ref{derivataRosselli} the MAE behaviour for different ML algorithms in the two -datasets (SNAQ and ENEA Pirelli) for the test datasets. We can clearly see that the dynamic GPR results are far better than the static GPR all along the derivative axis . In contrast, no improvement was observed with dynamic SVRs relatively to the static SVR (Fig. \ref{derivataRosselli}) due to the absence of sensors dynamic related information in the ENEA Pirelli dataset. This analysis helps clarify that the amelioration is not due to time series prediction capabilities of tapped delay architectures. In figures \ref{sensoriFonollosa} and \ref{tdnn_fonollosa5e60} we show how TDNN output responds significantly faster to abrupt transients, occurring in the UCSD dataset, with respect to raw sensor data, while figure \ref{comparisonNN} depicts  TDNN similar advantage with respect to the static NN algorithm . These results also show that this may sometimes occur at the cost of a limited observed over elongation. The enhanced error performances are, hence, not due to noise suppression capabilities of tapped delay architectures but to an improved responsiveness. 
Those figures, in facts, confirms at a wider level what already shown in Esposito et al. only for NNs \cite{esposito2016dynamic} using a single dataset. In that work, authors firstly shown that the MAE grows along with the speed with which the concentration changes but dynamic calibration algorithms can reduce the error notwithstanding the rapidity of change. Quite often, the faster the concentration transient, the better the improvement made. From the results of the three datasets, we infer that dynamic algorithms improve global performances primarily by reducing the error associated with rapid transients in static algorithms. 

\newpage

\begin{figure}[H]
	\centering
	\includegraphics[height=250pt,width=420pt]{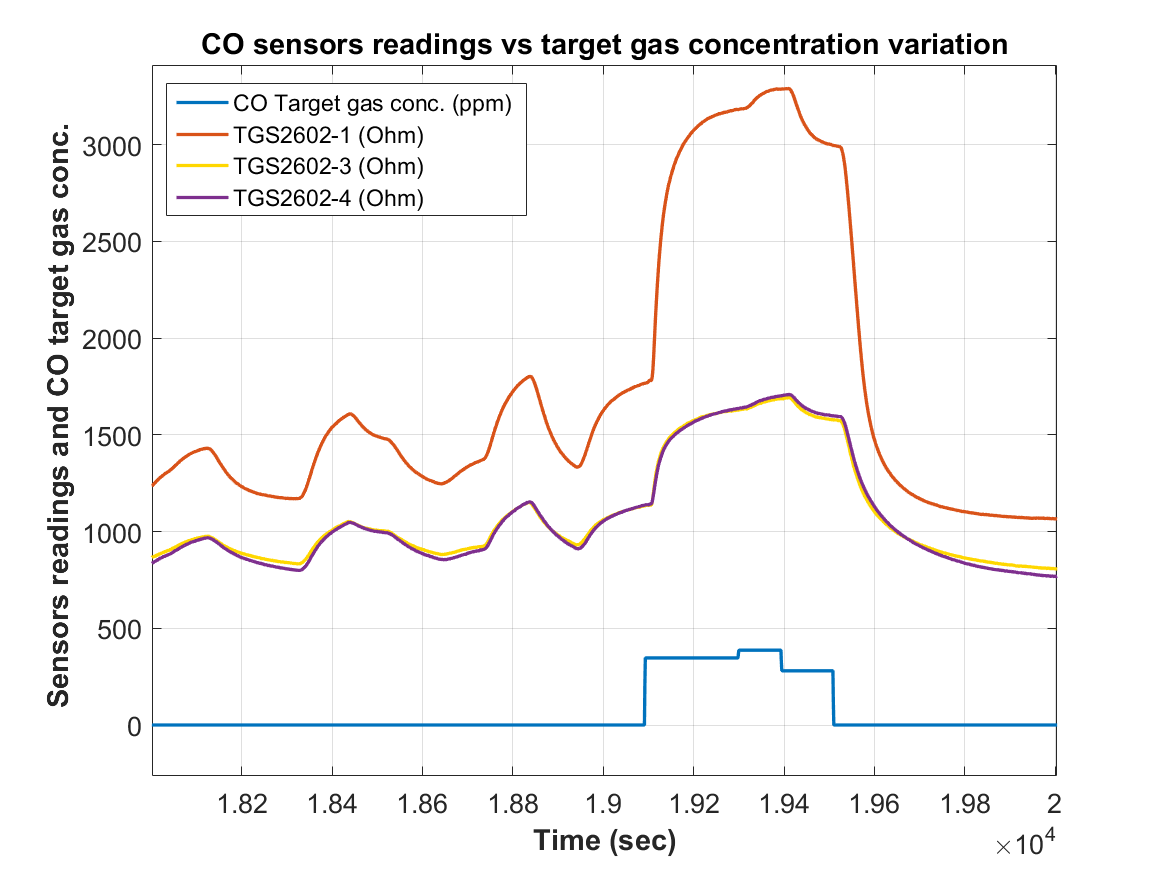}
\end{figure}

\clearpage 
\section{Conclusions}
Recently, chemical multi-sensor devices are increasingly designed for pervasive or mobile air quality deployments, often requiring computational intelligence to effectively solve their complex calibration problem. In this work, we have assessed and compared, for the first time, the performances of multiple machine learning approaches in a comprehensive set of continuous and open sampling scenarios. Five of the best performing machine learning approaches in the recent chemical sensing literature, along with their dynamic implementations were reviewed and assessed. The tests were carried out by using three different  datasets spanning a significant variety of conditions designed to challenge the technological limits of chemical sensors eventuallly determining the absolute performance levels. From the dynamic point of view, the datasets included   smooth or challenging abrupt concentration variations scenarios. Each of the sensors, used in the concerned devices, had its own dynamic behaviour versus the target gas as shown in their datasheet reported properties. In fact, data are sampled or averaged at different time resolution ranging from $1hr$ to $1/100th$ of second. These datasets were generated using devices that employed  different sensor technologies while being exposed to typical outdoor and indoor pollutants, in uncontrolled field measurement as well as in laboratory tests, with unknown or known interferents at high concentrations. 
 The utilisation of external test sets, the extensive model selection and performance computing approach (encompassing the use of hundreds or thousands of hyperparameters combinations), helped to guarantee a fair comparison among the techniques, ensuring statistical consistency to the entire framework. The obtained results, to our best knowledge, are comparable to, or ameliorate, the state of the art performances obtained for solid state based field calibrated air quality analyzers. 

The most relevant finding has been the consistency with which dynamic machine learning approaches surpasses their respective static counterparts that rely only on instantaneous sensor responses. This was observed in the two datasets with fast sampling period, suggesting a relevant impact of the slow sensor dynamic on the performances. Instead, none of the dynamic techniques shown a significant advantage when dealing with ENEA Pirelli dataset which features a 1hr averaging of sensor responses. Interestingly, the minimum effective length of the tapped delay line was found to be very similar among the different techniques. This confirms that these architectures are ideal for analysing, tracking and above all, correcting the intrinsic slow dynamic of the sensors, given a sufficient and optimal observation window length. Results also suggest to take into account the slowest sensor dynamic when designing tapped delay length in order to obtain a concise but effective representation. In both the field datasets, our results also indicated that a training set length of more than one week is required for this non-adaptive approaches to build a sufficient knowledge of the sensor array model.  In terms of basic machine learning techniques, SVRs was shown to have the best performance in most scenarios irrespective of the timeframe, the sensor technology or the length of the tapped delay line. However, despite not showing the best performance in many cases, plain shallow neural networks (NN) provided more compact and low computational/storage impact models for a usually very small performance cost. Non linear techniques shown a significant performance advantage over plain MLR being also able to use less training samples for obtaining the same performance levels. RC showed average performance levels needing small computational efforts for training at the cost of a very redundant model. Eventually, these results strongly suggest the use of dynamic approaches for on-line processing of chemical sensor data instead of the traditional static approach, especially when a network or mobile deployment is concerned. We observed that of all the models considered in this work, shallow neural networks confirms an interesting suitability for this task. Notwithstanding being one of the oldest methodology, they clearly prove to be one of the first choice that will be recommended for the next generation of intelligent, pervasive or wearable sensing systems with on board pollutant concentration estimation capabilities.
\section{Acknowledgments}
The authors want to thank the UCSD team for having shared their dataset used in this work. Equally, We want to thank the authors of \cite{oger2012} and \cite{fonollosa2015} for having shared, respectively, the basic framework and specific code used for developing our own Reservoir Computing implementation.
\section{Author Contributions}
S. De Vito and E. Esposito devised the work and designed the experimental optimization and testing framework. Together with M. Salvato they implemented the machine learning components and their optimization and performance assessment  code.  O. Popoola and R. Jones recorded, preprocessed and reviewed the inception of the SNAQ dataset. F. Formisano reviewed the final code and managed the parallel code execution. S. De Vito, E. Esposito and O. Popoola wrote the paper. S. De Vito and G. Di Francia supervised the entire work.

\newpage
\bibliographystyle{unsrt}
\bibliography{biblioRevised} 
\end{document}